\documentclass[journal]{IEEEtran}
\usepackage{cite}

\ifCLASSINFOpdf
\else
\fi

\usepackage{epsfig}
\usepackage{graphicx, float, caption, subcaption, url}
\usepackage{amsmath}
\usepackage{amssymb}
\usepackage{amsfonts}
\usepackage{multirow}
\usepackage{booktabs}

\begin{document}
%
\title{Privacy Leakage of SIFT Features via Deep Generative Model based Image Reconstruction}

%
%
%

\author{Haiwei~Wu, \IEEEmembership{Student~Member,~IEEE} and~Jiantao~Zhou,~\IEEEmembership{Senior~Member,~IEEE}
\thanks{The authors are with the State Key Laboratory of Internet of Things for Smart City, and also with the Department of Computer and Information Science, Faculty of Science and Technology, University of Macau, Macau 999078, China (\emph{Corresponding Author: Jiantao Zhou, email: jtzhou@umac.mo}).}
}

%
%

\markboth{IEEE Transactions on Information Forensics and Security}%
{SIFT-based Image Reconstruction via Different Prior Knowledge}
%




\maketitle

\begin{abstract}

Many practical applications, e.g., content based image retrieval and object recognition, heavily rely on the local features extracted from the query image. As these local features are usually exposed to untrustworthy parties, the privacy leakage problem of image local features has received increasing attention in recent years. In this work, we thoroughly evaluate the privacy leakage of Scale Invariant Feature Transform (SIFT), which is one of the most widely-used image local features. We first consider the case that the adversary can fully access the SIFT features, i.e., both the SIFT descriptors and the coordinates are available. We propose  a novel end-to-end, coarse-to-fine deep generative model for reconstructing the latent image from its SIFT features. The designed deep generative model consists of two networks, where the first one attempts to learn the structural information of the latent image by transforming from SIFT features to Local Binary Pattern (LBP) features, while the second one aims to reconstruct the pixel values guided by the learned LBP. Compared with the state-of-the-art algorithms, the proposed deep generative model produces much improved reconstructed results over three public datasets. Furthermore, we address more challenging cases that only partial SIFT features (either SIFT descriptors or coordinates) are accessible to the adversary. It is shown that, if the adversary can only have access to the SIFT descriptors while not their coordinates, then the modest success of reconstructing the latent image can be achieved for highly-structured images (e.g., faces) and would fail in general settings. In addition, the latent image can be reconstructed with reasonably good quality solely from the SIFT coordinates. Our results would suggest that the privacy leakage problem can be largely avoided if the SIFT coordinates can be well protected.


\end{abstract}


%
\IEEEpeerreviewmaketitle

\section{Introduction}\label{sec:introduction}
%
%
%
%

As one of the most popular algorithms in computer vision to extract and encode local features, Scale Invariant Feature Transform (SIFT) \cite{lowe2004distinctive} has been proven to be very robust against various distortions \cite{mikolajczyk2005performance, qin2014towards} and has been widely employed in many practical scenarios, e.g., content based image retrieval (CBIR) \cite{amsaleg2001cbir, torres2006cbir2, torres2009cbir3}, object recognition \cite{rothganger2006object}, visual tracking \cite{gauglitz2011evaluation}, and image matching \cite{li2019fast}. Due to its extreme popularity, the privacy and security issues regarding the SIFT features have been attracting increasing attention. For instance, in our recent studies, it was demonstrated that SIFT keypoints can be maliciously removed and forged with negligible distortions on the original image, making the decisions from SIFT-based systems untrustworthy \cite{li2019fast, li2017removal}.

\begin{figure}[t!]
	\begin{subfigure}{.115\textwidth}
		\centering
		\includegraphics[width=\textwidth]{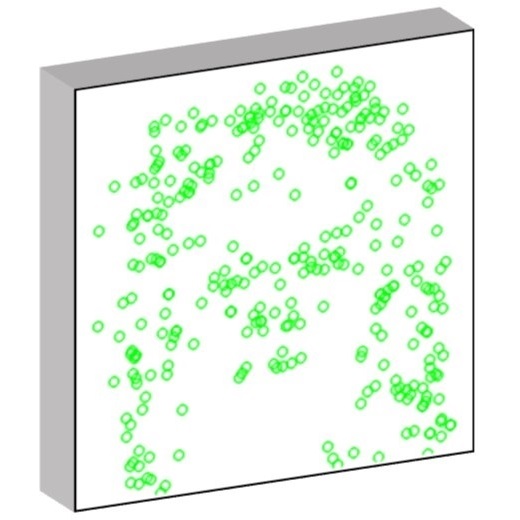}
	\end{subfigure}
	\begin{subfigure}{.115\textwidth}
		\centering
		\includegraphics[width=\textwidth]{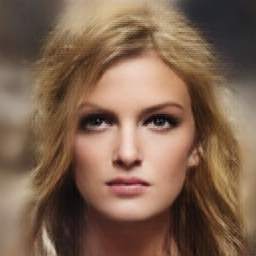}
	\end{subfigure}
	\begin{subfigure}{.115\textwidth}
		\centering
		\includegraphics[width=\textwidth]{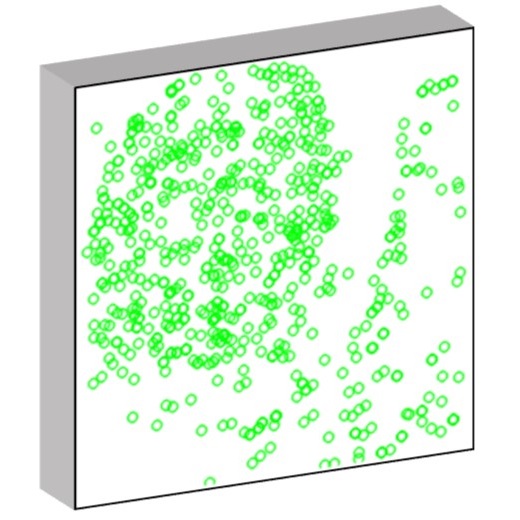}
	\end{subfigure}
	\begin{subfigure}{.115\textwidth}
		\centering
		\includegraphics[width=\textwidth]{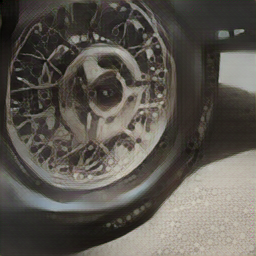}
	\end{subfigure}
	\newline
	\newline
	\newline
	\begin{subfigure}{.115\textwidth}
		\centering
		\includegraphics[width=\textwidth]{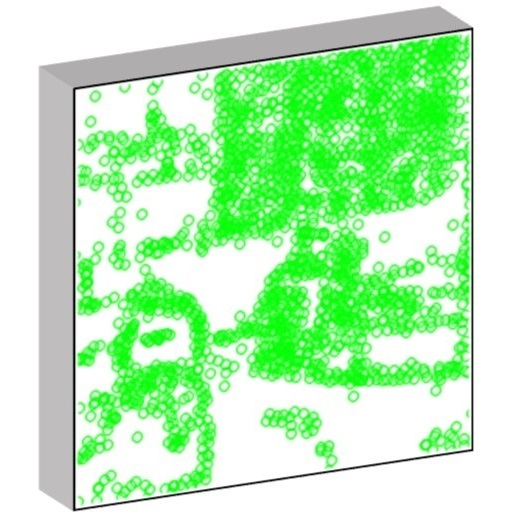}
	\end{subfigure}
	\begin{subfigure}{.115\textwidth}
		\centering
		\includegraphics[width=\textwidth]{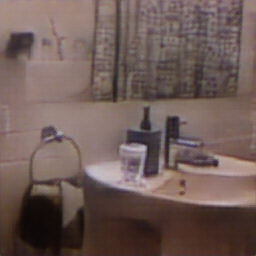}
	\end{subfigure}
	\begin{subfigure}{.115\textwidth}
		\centering
		\includegraphics[width=\textwidth]{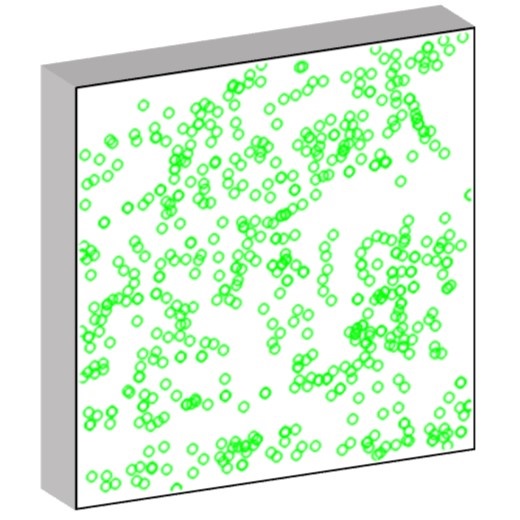}
	\end{subfigure}
	\begin{subfigure}{.115\textwidth}
		\centering
		\includegraphics[width=\textwidth]{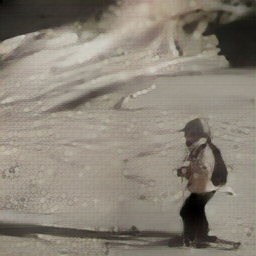}
	\end{subfigure}
	\caption{Reconstruction results of our proposed model on images of face, wheel, indoor and outdoor scenes. In each pair, the left is the input SIFT features and the right is the reconstructed image.}
	\label{fig:heading_figure}
\end{figure}

Noticing that SIFT features are often exposed to untrustworthy parties, we in this work thoroughly evaluate the privacy leakage problem of SIFT features. More specifically, we consider the following two scenarios, where full or partial SIFT features can be accessed by an adversary:

\begin{itemize}
\item \textbf{Scenario I}: Both the SIFT descriptors and their coordinates are accessible to the adversary. For instance, in 3D point clouds based systems \cite{irschara2009sfm, sattler2011fast, li2012worldwide, lim2015real}, 3D object recognition \cite{lowe2001object_recog} and panoramic image stitching \cite{brown2007automatic}, users need to provide both the SIFT descriptors \emph{and} the coordinates, potentially leaking them out.
\item \textbf{Scenario II}: Only SIFT descriptors \emph{or} their coordinates are accessible to the adversary. For instance, in many CBIR systems \cite{amsaleg2001cbir, torres2006cbir2, torres2009cbir3} and copy-move forgery detection systems \cite{pan2010copymove, amerini2011copymove, li2019fast}, it is sufficient to only provide the SIFT descriptors, while not their coordinates.
\end{itemize}


In order to evaluate the risk of the information leakage from SIFT features, we need to know how much information is carried by them. A feasible solution to this question is to investigate to what extent the latent image can be recovered from these SIFT features or local features in general.  Along this line, several approaches \cite{weinzaepfel2011reconstructing, angelo2012beyond, vondrick2013hoggles, kato2014image, desolneux2017stochastic, mahendran2015understanding} have been devised to reconstruct the images from local features, mainly under the assumption of \textbf{Scenario I}, i.e., full features can be accessed. The pioneer study was conducted by Weinzaepfel \textit{et al.} \cite{weinzaepfel2011reconstructing},  who attempted to reconstruct the image from SIFT features through patch searching, pasting and smoothing. However, due to the sparse nature of local descriptors, only some rough contours can be recovered, while the fine textures are missing. Angelo \textit{et al.} \cite{angelo2012beyond} later proposed an inverse optimization framework for recovering the latent image from the local binary descriptors, without relying on any external databases. Vondrick \textit{et al.} \cite{vondrick2013hoggles} addressed the problem of the image reconstruction from the histograms of gradient orientations (HOG) descriptors by using the dictionary representation. Through estimating the spatial arrangement of local descriptors over a large-scale image database, Kato \textit{et al.} \cite{kato2014image} presented a method to reconstruct the image from its Bag-of-Visual-Words (BoVW) feature. More recently, Desolneux \textit{et al.} \cite{desolneux2017stochastic} devised two reconstruction models for HOG features by adopting the Poisson editing, capable of recovering global shapes and many geometric details. To further improve the reconstruction performance, there is a recent trend of using deep convolutional neural networks (CNNs) and generative adversarial networks (GANs) \cite{dosovitskiy2016inverting, mahendran2015understanding, wu2019image, pittaluga2019revealing}.  Dosovitskiy and Brox \cite{dosovitskiy2016inverting} proposed a reconstruction approach from  local features through an encoder-decoder neural network. Wu \textit{et al.} \cite{wu2019image} then improved it by introducing GANs architecture and a multi-scale features generation. Further, Pittaluga \textit{et al.} \cite{pittaluga2019revealing} trained a cascade of U-Nets with extra convolutional layers to reveal scenes from the local features. Unfortunately, these methods tend to generate severe boundary artifacts and distorted structures. 

In this work, we first consider the case that the adversary can fully access the SIFT features (both descriptors and coordinates), i.e., under \textbf{Scenario I}. We thoroughly evaluate the privacy leakage of SIFT features by constructing a novel end-to-end, coarse-to-fine image reconstruction model, SIFT-LBP-Image (SLI), that consists of two networks. The first network, called LBP reconstruction network, attempts to learn the structural information of the latent image by transforming from SIFT features to LBP features, while the second one aims to reconstruct the pixel values guided by the learned LBP. Extensive experiments on three publicly available datasets \texttt{CelebA-HQ} \cite{karras2017progressive}, \texttt{MD-NYU} \cite{pittaluga2019revealing} and \texttt{ImageNet} \cite{deng2009imagenet} demonstrate that our proposed model can generate better results than the state-of-the-art competitors, both quantitatively and qualitatively (see Fig. \ref{fig:heading_figure} for some examples). Furthermore, we address more challenging cases where only partial SIFT features are available, i.e., under \textbf{Scenario II}. In the case that the SIFT coordinates are not accessible, we design two methods for predicting the missing coordinate information, which achieve modest success for highly-structured images (e.g., faces), while would fail for general settings (e.g., buildings).  The challenge mainly comes from the fact that, for general cases, there is no strong correlation between the descriptor and its absolute coordinate, i.e., the extracted descriptor could be the same regardless the location of the keypoint. We also evaluate the possibility of reconstructing the latent image solely from the coordinates. It is found that the rough contour of the latent image can still be reconstructed, though the fine textures are missing. Our results would suggest that the coordinates play a more critical role in ensuring the privacy of the SIFT features. In other words, if the coordinates of the SIFT features can be well protected, the sensitive information leakage can be largely avoided.



Our major contributions can be summarized as follows:

\begin{itemize}
\item We propose SLI, an end-to-end, coarse-to-fine deep generative model to recover the latent image from its SIFT features.


\item Our model SLI achieves better reconstruction performance in comparison with several state-of-the-art methods \cite{desolneux2017stochastic, dosovitskiy2016inverting, pittaluga2019revealing} over a variety of challenging datasets including \texttt{CelebA-HQ} \cite{karras2017progressive}, \texttt{MD-NYU} \cite{pittaluga2019revealing} and \texttt{ImageNet} \cite{deng2009imagenet}.

\item We investigate the challenging cases where the adversary can only access partial SIFT features (either descriptors or coordinates). To the best of our knowledge, it is the first work to specifically address the problem of reconstructing the latent image from the incomplete SIFT features. We demonstrate that the reconstruction performance is greatly degraded when coordinates are missing, especially for those images without regular structures.

\end{itemize}

The rest of this paper is organized as follows. Section \ref{sec:related_works} briefly reviews the SIFT and LBP algorithms. Section \ref{sec:methods} presents our proposed model SLI under \textbf{Scenario I} and Section \ref{sec:advanced_methods} introduces the reconstruction approaches under \textbf{Scenario II}. Extensive experiments are then given in Section \ref{sec:experiments}, and finally Section \ref{sec:conclusion} concludes.


\section{Introduction of SIFT and LBP}\label{sec:related_works}

In this section, we provide a brief introduction of SIFT and LBP algorithms.

\subsection{SIFT Features Generation and Matching}
The detection of SIFT keypoints and the generation of their corresponding descriptors can be roughly divided into four steps: i) establishment of scale space; ii) detection and refinement of extreme points; iii) assignment of dominant orientation; and iv) generation of descriptors.

At step i), by repeatedly convolving an input image $\mathbf{I}$ with Gaussian filters at different scales, the Gaussian-blurred image $L(x,y, \sigma)$ can be computed as
\begin{equation}
L(x, y, \sigma) = \mathbf{I}(x, y) \otimes G(x, y, \sigma).
\end{equation}
Here $G(x, y, \sigma)$ is the Gaussian kernel at scale $\sigma$, i.e.,
\begin{equation}
G(x, y, \sigma) = \frac{1}{2\pi\sigma^2}\mathrm{e}^{-(x^2+y^2)/2\sigma^2}
\end{equation}

At step ii), a series of candidate SIFT keypoints are detected from the local extrema within a $3 \times 3 \times 3$ cube of the Difference of Gaussians (DoG) domain, where the DoG image at scale $\sigma$ is calculated by the difference of adjacent Gaussian-blurred images
\begin{equation}
D(x, y, \sigma) = L(x, y, k\sigma) - L(x, y, \sigma),
\end{equation}
where $k$ is a predefined constant. In order to reject unstable extreme points in the DoG domain, a contrast threshold and an edge threshold are used for keypoints refinement.


At step iii), the orientation of each point $(x, y, \sigma)$ is defined as
\begin{equation}
\theta(x, y, \sigma) = {\rm tan}^{-1}(\frac{d_y}{d_x}),
\end{equation}
where $d_x$ and $d_y$ are the horizontal and vertical gradients of $(x, y, \sigma)$. An orientation histogram is constructed by gathering the orientation of points in a local window centered at the SIFT keypoint. The maximum value in the orientation histogram is assigned as the dominant orientation to guarantee the rotation invariance.

At step iv), a 128-dimensional descriptor $\mathbf{f}$ is calculated from the gradient information of 8 directions in a $16 \times 16$ local area centered at the SIFT keypoint.

Through the above four steps, for the image $\mathbf{I}$, we can generate a list of $n$ keypoints $\mathcal{K} = \{\mathbf{k}_1, \mathbf{k}_2, \cdots, \mathbf{k}_n\}$ and their corresponding descriptors $\mathcal{F} = \{\mathbf{f}_1, \mathbf{f}_2, \cdots, \mathbf{f}_n\}$. Specifically, each SIFT keypoint $\mathbf{k}$ is a four-dimensional vector
\begin{equation}
\mathbf{k} = (x, y, \sigma, \theta),
\end{equation}
where $(x, y)$ denotes the coordinate of the SIFT keypoint in the image plane, $\sigma$ and $\theta$ represent the scale and dominant orientation, respectively. For a given image, its SIFT features are composed of two parts $\mathcal{K}$ and $\mathcal{F}$.
%


Upon having the SIFT features, a SIFT keypoints matching algorithm was also suggested in \cite{lowe2004distinctive}. Specifically, let $\mathbf{d}=\{d_1, d_2, \cdots, d_{n-1}\}$ record the Euclidean distances between the descriptor $\mathbf{f}_i$ and the remaining descriptors $\{\mathbf{f}_j \}$ ($j \neq i$) in an increasing order, i.e., $d_1 \leq d_2 \leq \cdots \leq d_{n-1}$. Then a pair of reliable SIFT match exists if and only if
\begin{equation}
d_1 / d_2 < t,
\end{equation}
where $t \in (0, 1)$ is a predefined parameter commonly set as 0.8.

\begin{figure}[t!]
	\centering
	\includegraphics[width=.485\textwidth]{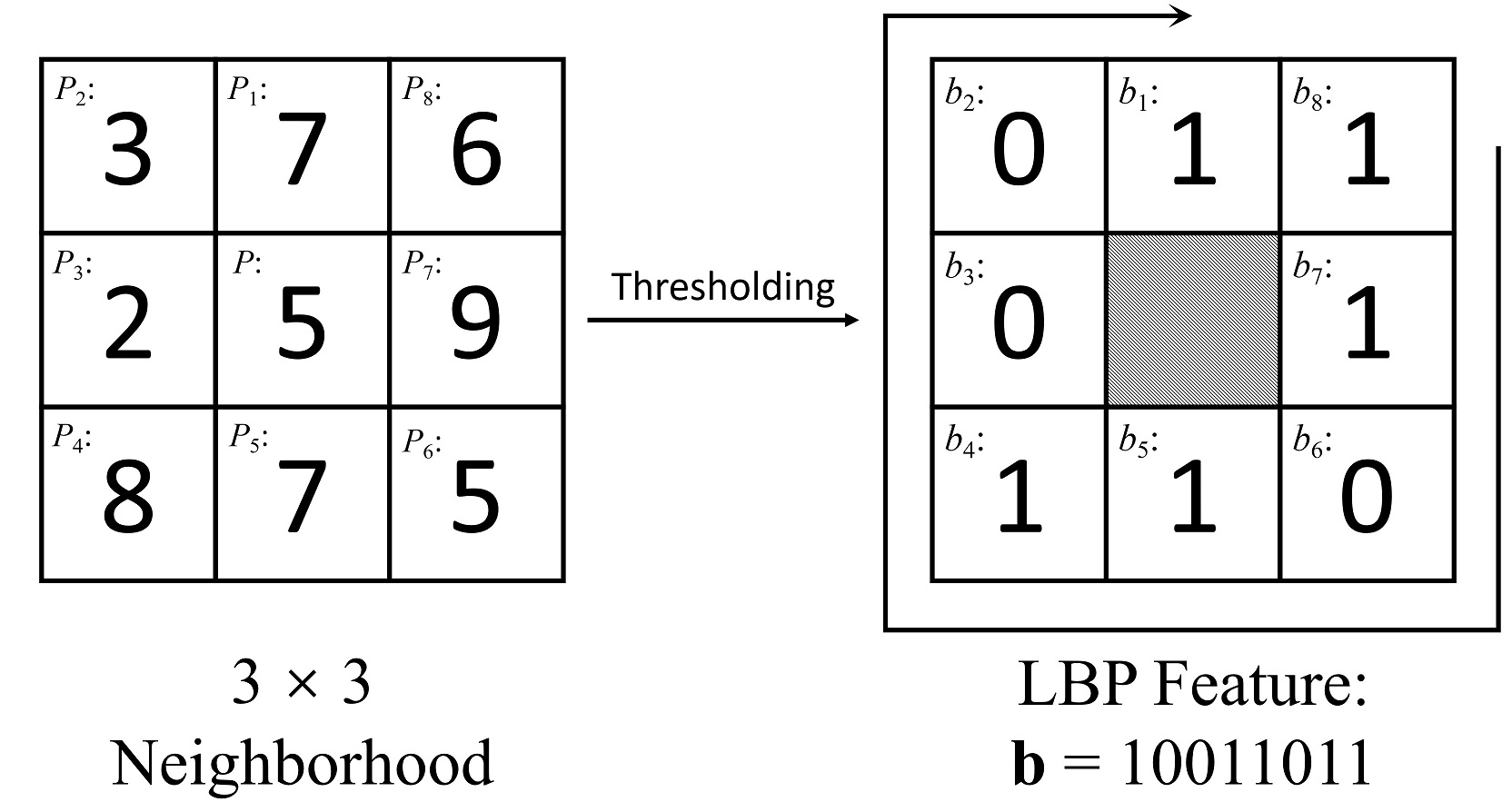}
	\caption{An example of the LBP extraction. Left is the original $3 \times 3$ neighborhood. Right is the thresholded neighborhood, and the LBP feature of the centering pixel $P$ is $\mathbf{b} = 10011011$.}
	\label{fig:LBP_extraction}
\end{figure}

\begin{figure*}[t!]
	\centering
	\includegraphics[width=\textwidth]{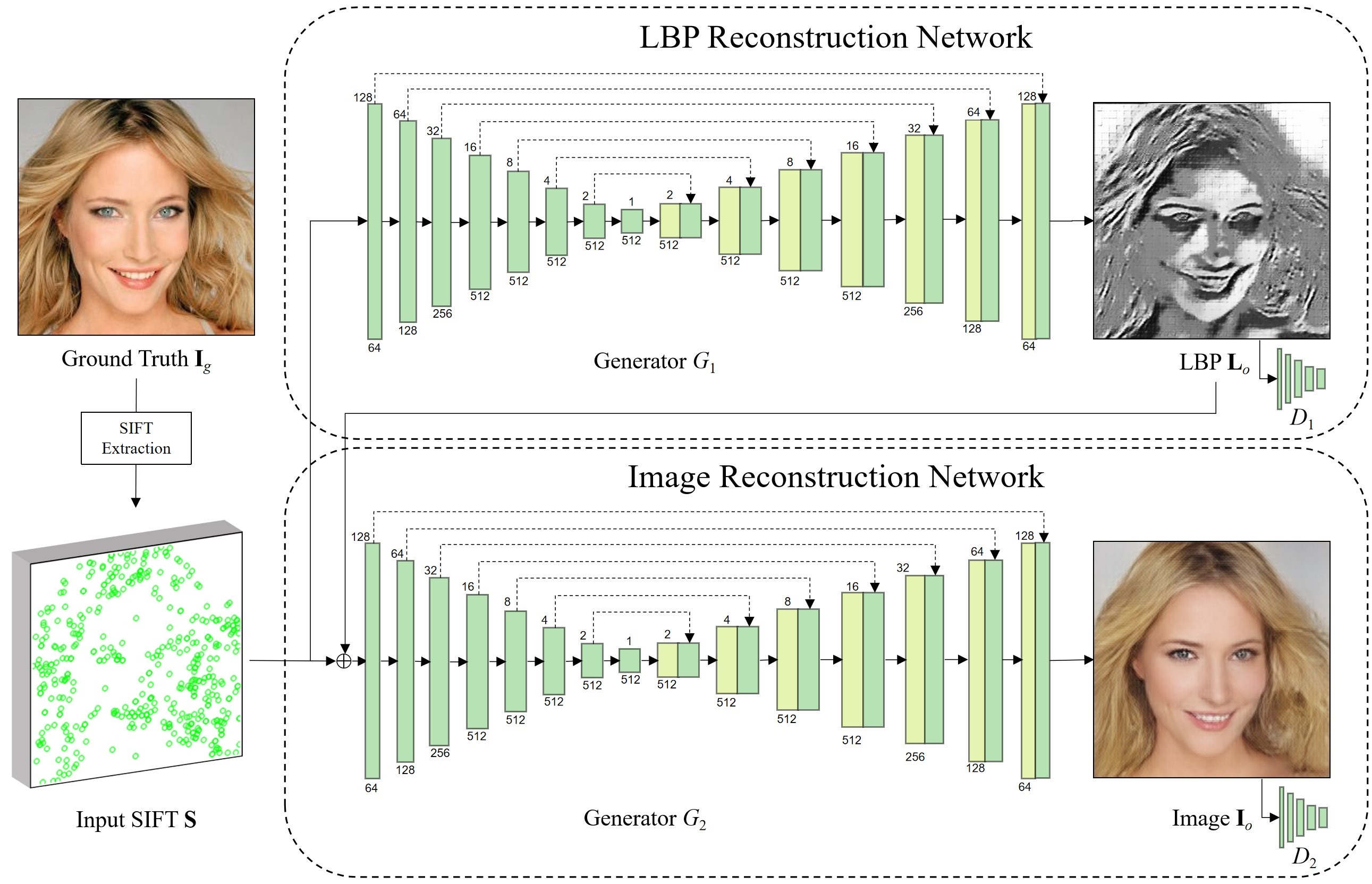}
	\caption{Overview of our proposed SIFT-LBP-Image (SLI) reconstruction model. The number above each layer represents the size of the resolution, while the number below means the dimension.}
	\label{fig:framework}
\end{figure*}

\subsection{Local Binary Pattern (LBP)}
LBP is a widely used texture descriptor originally proposed by Ojala \cite{ojala1996comparative}. The LBP features extraction process is to label each pixel of an image by thresholding its spatial neighborhood. Specifically, to extract the LBP features associated with the pixel $P$, we first obtain its $M\times N$ neighborhood denoted by $P_1, P_2, \cdots, P_{MN-1}$. Then the LBP features associated with $P$ is a string of binary bits $\mathbf{b} = b_1, b_2, \cdots, b_{MN-1}$. where

\begin{equation}
b_i=
\begin{cases}
0& \mathrm{if}~ P_i \leq P\\
1& \mathrm{otherwise}
\end{cases},
\mathrm{for}~i = 1, \cdots, MN-1.
\end{equation} An example of the LBP features extraction is illustrated in Fig. \ref{fig:LBP_extraction}, where $M=N=3$.

LBP features essentially record the relative ordering within a block of pixels, capturing the information of edges, spots and other local structures \cite{zhang2010local}. LBP shows very good performance in many vision tasks, e.g., unsupervised texture segmentation \cite{ojala1999unsupervised}, face recognition \cite{ahonen2006face}, and image reconstruction \cite{waller2013image}.


\section{Image Reconstruction from Full SIFT Features}\label{sec:methods}

In this section, we consider the problem of reconstructing the latent image under \textbf{Scenario I}, i.e., full SIFT features are accessible to the adversary. We first present the architecture of the proposed SIFT-LBP-Image (SLI) deep generative model, and then give the details on the model optimization. We experimentally find that the scale $\sigma$ and the dominant orientation $\theta$ only bring negligible reconstruction performance gains, and hence, they are abandoned. In other words, the SIFT descriptors $\mathcal{F}$ and the associated coordinates $(x, y)$  are used as the features map to be injected into SLI.

\subsection{SIFT-LBP-Image (SLI) Model}

The architecture of the proposed SIFT-LBP-Image (SLI) model is illustrated in Fig. \ref{fig:framework}. As can be seen, SLI is an end-to-end, coarse-to-fine deep generative model, consisting of two networks. The first one called LBP reconstruction network transforms the SIFT features into LBP features, providing structural information to assist the subsequent image reconstruction network, which aims to complete the actual image reconstruction task. One of the reasons why we select LBP features under this circumstance is that it contains a great amount of structural information, capable of well guiding the image reconstruction task. As verified in \cite{waller2013image}, an image visually close to the original one could be reconstructed solely from its LBP features. Also, from the perspective of practical implementation, LBP is easy to be computed and very few parameters are involved. More importantly, as expected and will be verified experimentally, the conversion from SIFT features to LBP, and eventually to image significantly improves the reconstruction performance, compared with the challenging task of reconstructing the latent image directly from its SIFT features.


Both networks follow an adversarial model \cite{goodfellow2014generative}, i.e., each network contains a generator based on U-Net architecture \cite{ronneberger2015u}, and a discriminator based on the PatchGAN \cite{isola2017image}. Let $\mathcal{K}$ and $\mathcal{F}$ be the SIFT keypoints and descriptors extracted from an input image $\mathbf{I} \in \mathbb{R}^{H \times W \times 3}$ in the grayscale channel. Denote $\mathbf{S} \in \mathbb{R}^{H \times W \times 128}$ as the input SIFT features map, where descriptors $\mathcal{F}$ are assigned to their corresponding coordinates and zero vectors elsewhere. At the training stage, the generator of the LBP reconstruction network $G_1 : \mathbb{R}^{H \times W \times 128} \to \mathbb{R}^{H \times W \times 1}$ takes $\mathbf{S}$ as input, and outputs the estimated LBP $\mathbf{L}_o$. During this process, the discriminator $D_1 : \mathbb{R}^{H \times W \times 1} \to \mathbb{R}$ works together with the $G_1$ to produce the result $\mathbf{L}_o$. Upon having a well-estimated LBP, we then use it to guide the image reconstruction process in the subsequent network. Specifically, the generator $G_2 : \mathbb{R}^{H \times W \times (128+1)} \to \mathbb{R}^{H \times W \times 3}$ takes $(\mathbf{S}$, $\mathbf{L}_o)$ as input, and outputs the final reconstructed result $\mathbf{I}_o$, with the assistance of the discriminator $D_2 : \mathbb{R}^{H \times W \times 3} \to \mathbb{R}$. At the testing stage, the procedure is similar, but without the need of using the two discriminators $D_1$ and $D_2$.

For the $G_1$ or $G_2$, we adopt a pruned U-Net architecture \cite{ronneberger2015u} composed of an encoder and a decoder. In the encoder, each layer has a $4\times4$ convolution, an Instance Norm \cite{ulyanov2016instance} and a LeakyReLU \cite{xu2015empirical} with $\alpha=0.2$. The decoder has a symmetric structure, except that the convolution and LeakyReLU are replaced with the deconvolution and ReLU \cite{nair2010rectified}, respectively. Additionally, skip connections are used to concatenate the features from each layer of the encoder with the corresponding layer of the decoder. Experimentally, we find that the dilated convolutions in the original U-Net architecture \cite{ronneberger2015u} bring negligible improvements to the final reconstruction results. We hence prune the U-Net architecture by removing the dilated convolutions, so as to reduce the number of model parameters, which could speed up the training process. For the $D_1$ or $D_2$, we adopt the PatchGAN architecture \cite{isola2017image}.

\subsection{Optimization of the proposed networks}

For the optimization of the LBP reconstruction network, we use the combination of $\ell_1$ reconstruction loss \cite{ledig2017photo}, $\ell_2$ perceptual loss \cite{dosovitskiy2016perceptual} and adversarial loss \cite{martineau2018rsgan}. More specifically, the reconstruction loss is naturally defined as:
\begin{equation}\label{loss:recon}
\mathcal{L}_{r} = ||\mathbf{L}_{o} - \mathbf{L}_{g}||.
\end{equation}
The perceptual loss penalizes the reconstructed LBP that is not perceptually similar to the ground-truth LBP $\mathbf{L}_g$, and it can be defined as:
\begin{equation}\label{loss:perc}
\mathcal{L}_p=\sum_{h\in \mathcal{A}}||\varphi_h([\mathbf{L}_{o}, \mathbf{L}_{o}, \mathbf{L}_{o}])-\varphi_h([\mathbf{L}_{g}, \mathbf{L}_{g}, \mathbf{L}_{g}])||_2,
\end{equation}
where $\varphi_h$ is the activation map corresponding to the $h$-th layer of an ImageNet-pretrained VGG-16 network. The set $\mathcal{A}$ is formed by the layer indexes of $\rm conv2\_1$, $\rm conv3\_1$, $\rm conv4\_1$ layers. Here we concatenate three $\mathbf{L}_{o}$ or $\mathbf{L}_{g}$ as the input of layers in set $\mathcal{A}$ because VGG-16 fixes the input as three channels. Also, the Relativistic average GAN (RaGAN) \cite{martineau2018rsgan} can be calculated as follows:
\begin{equation}\label{loss:adv}
\mathcal{L}_{D_1} = -\mathbb{E}_{\mathbf{L}_g}\big[\mbox{log}\big(\widetilde{D}(\mathbf{L}_g)\big)\big]-\mathbb{E}_{\mathbf{L}_o}\big[\mbox{log}\big(1-\widetilde{D}(\mathbf{L}_o)\big)\big],
\end{equation}
\begin{equation}
\mathcal{L}_{G_1} = -\mathbb{E}_{\mathbf{L}_o}\big[\mbox{log}\big(\widetilde{D}(\mathbf{L}_o)\big)\big]-\mathbb{E}_{\mathbf{L}_g}\big[\mbox{log}\big(1-\widetilde{D}(\mathbf{L}_g)\big)\big],
\end{equation}
where
\begin{equation}
\widetilde{D}(\mathbf{L}_g) = \mbox{sigmoid}\big(D_1(\mathbf{L}_g)-\mathbb{E}_{\mathbf{L}_o}[D_1(\mathbf{L}_o)]\big),
\end{equation}
\begin{equation}
\widetilde{D}(\mathbf{L}_o) = \mbox{sigmoid}\big(D_1(\mathbf{L}_o)-\mathbb{E}_{\mathbf{L}_g}[D_1(\mathbf{L}_g)]\big).
\end{equation}
Finally, the loss functions for the LBP reconstruction network are defined by integrating the above three types of loss:
\begin{equation}
\mathcal{L}^{LBP}_{G_1} = \lambda_r\mathcal{L}_r + \lambda_p\mathcal{L}_p + \lambda_{g}\mathcal{L}_{G_1},
\end{equation}
\begin{equation}\label{loss:dis}
\mathcal{L}^{LBP}_{D_1} = \mathcal{L}_{D_1},
\end{equation}
where $\lambda_r$, $\lambda_p$ and $\lambda_a$ are the parameters trading off different types of loss, whose settings will be clarified in Section \ref{sec:experiments}.

For the loss function of the image reconstruction network, we similarly adopt the combination of $\ell_1$ reconstruction loss, $\ell_2$ perceptual loss and adversarial loss. Besides, to better optimize the high-level features of the image reconstruction network, we further introduce the style loss \cite{gatys2016image}, which is used to measure the differences between the covariances of the activation maps. This is an effective strategy to eliminate the ``checkerboard'' artifacts caused by deconvolution layers \cite{sajjadi2017enhancenet}. Typically, the style loss can be defined as:
\begin{equation}
\mathcal{L}_s= \sum_{h\in \mathcal{A}}|| \mathbf{G}^{\varphi_h}(\mathbf{I}_{o}) - \mathbf{G}^{\varphi_h}(\mathbf{I}_{g}) ||_2,
\end{equation}
where $\mathbf{G}^{\varphi_h}$ is a $3 \times 3$ Gram matrix constructed from the activation map $\varphi_h$.

Finally, the loss functions for the image reconstruction network can be computed as:
\begin{equation}
\begin{aligned}
\mathcal{L}^{IMG}_{G_2} =& \lambda_s\mathcal{L}_s + \\
&\lambda_r||\mathbf{I}_{o} - \mathbf{I}_{g}|| +\\
&\lambda_p\sum_{h\in \mathcal{A}}||\varphi_h(\mathbf{I}_{o})-\varphi_h(\mathbf{I}_{g})||_2 - \\ &\lambda_{g}\big[\mathbb{E}_{\mathbf{I}_o}\big[\mbox{log}\big(\widetilde{D}(\mathbf{I}_o)\big)\big]+\mathbb{E}_{\mathbf{I}_g}\big[\mbox{log}\big(1-\widetilde{D}(\mathbf{I}_g)\big)\big]\big],
\end{aligned}
\end{equation}

\begin{equation}
\mathcal{L}^{IMG}_{D_2} = -\mathbb{E}_{\mathbf{I}_g}\big[\mbox{log}\big(\widetilde{D}(\mathbf{I}_g)\big)\big]-\mathbb{E}_{\mathbf{I}_o}\big[\mbox{log}\big(1-\widetilde{D}(\mathbf{I}_o)\big)\big],
\end{equation}
where
\begin{equation}
\widetilde{D}(\mathbf{I}_g) = \mbox{sigmoid}\big(D_2(\mathbf{I}_g)-\mathbb{E}_{\mathbf{I}_o}[D_2(\mathbf{I}_o)]\big),
\end{equation}
\begin{equation}
\widetilde{D}(\mathbf{I}_o) = \mbox{sigmoid}\big(D_2(\mathbf{I}_o)-\mathbb{E}_{\mathbf{I}_g}[D_2(\mathbf{I}_g)]\big).
\end{equation}

To stabilize the training process and alleviate the gradient vanishing problem, we first train the generator $G_1$ and the discriminator $D_1$ in the LBP network. Then we concatenate $G_1$ to the image reconstruction network, and perform an end-to-end training over $G_1$, $G_2$ and $D_2$ simultaneously. Here, Adam \cite{kingma2014adam} algorithm is adopted.

We would also like to emphasize that we need to have access to the full SIFT features (both descriptors and coordinates) to train and use the SLI deep model. As mentioned previously, in many practical applications such as CBIR, the assumption on the availability of the full SIFT features is not valid, namely, the adversary can only access partial SIFT features: either descriptors or coordinates. In the next Section, we will tackle this challenge of reconstructing the latent image from partial SIFT features.


\section{Image Reconstruction from Partial SIFT Features}\label{sec:advanced_methods}

Since the SIFT features of a given image consist of a set of descriptors and coordinates, we consider two cases of partial SIFT features, namely, 1) absence of coordinates and 2) absence of descriptors. In the following, we discuss the latent image reconstruction for these two cases separately.


\subsection{Absence of Coordinates}

\begin{figure*}[t!]
	\centering
	\begin{subfigure}{\textwidth}
		\includegraphics[width=\textwidth]{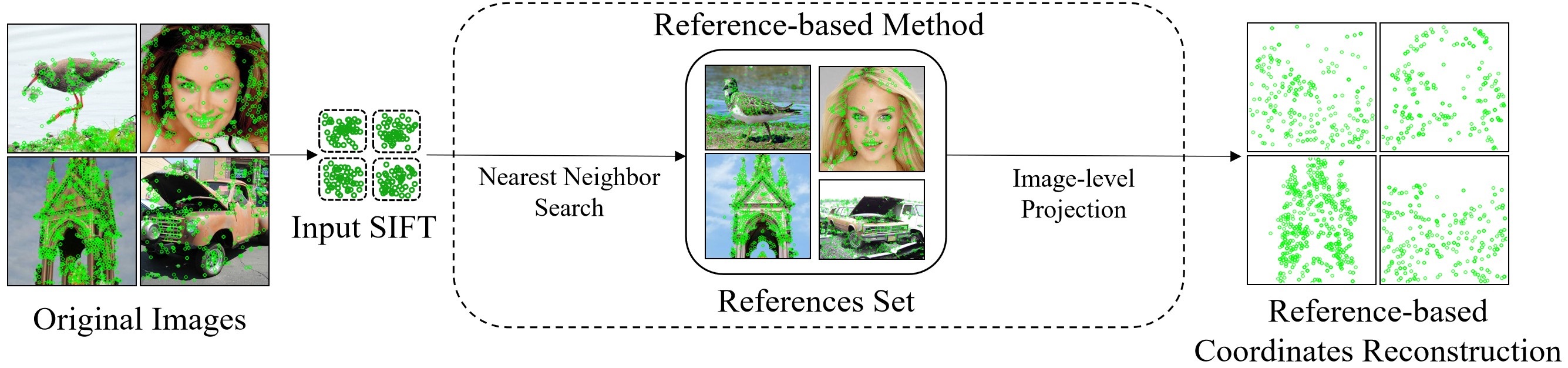}
		\caption{}
	\end{subfigure}
	
	\begin{subfigure}{\textwidth}
		\includegraphics[width=\textwidth]{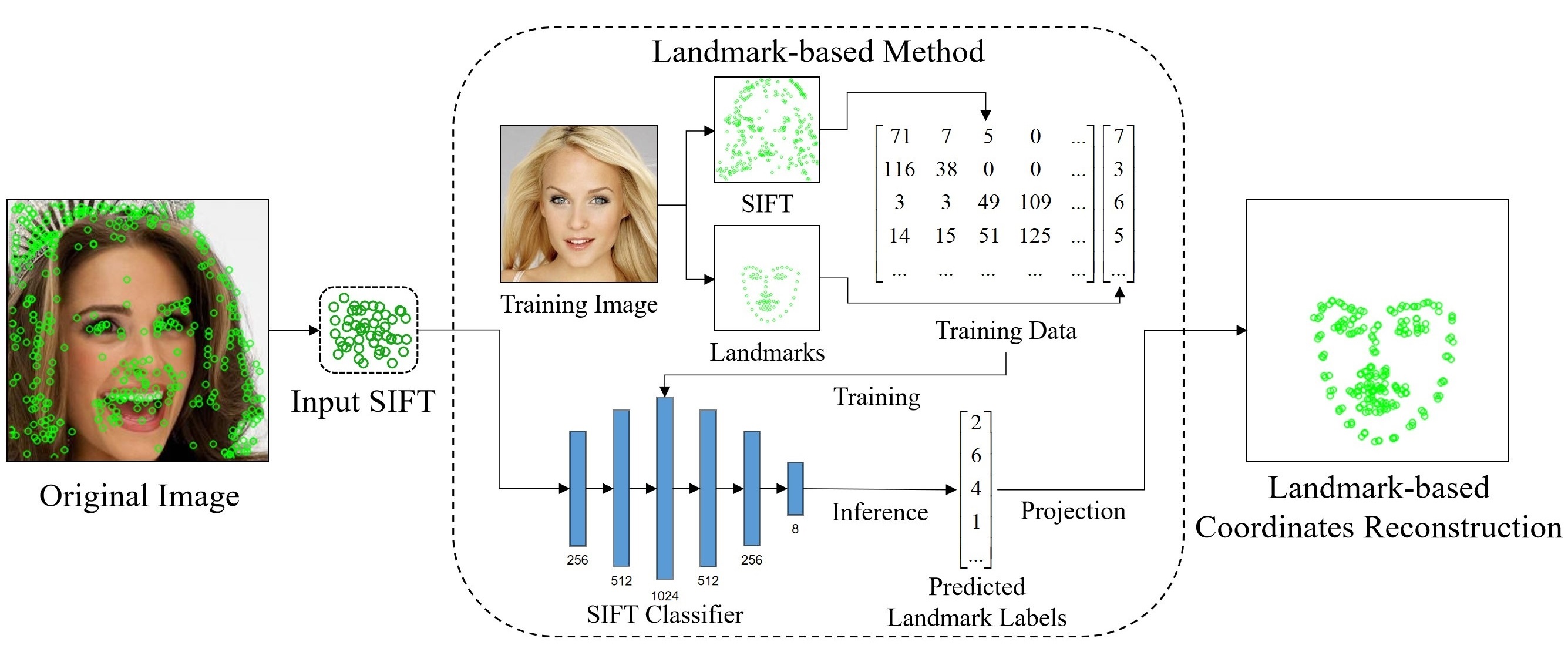}
		\caption{}
	\end{subfigure}
	\caption{Framework of reference-based (a) and landmark-based (b) methods for coordinates estimation. The number below each layer of the SIFT classifier means the dimension.}
	\label{fig:coordinates_reconstruction}
\end{figure*}

Clearly, SIFT descriptors without the corresponding coordinates cannot be directly used as model input of the proposed SLI presented in Section III. A natural solution to this problem is to somehow estimate the coordinates of these SIFT descriptors, and then the deep generative model SLI can be applied. It should be pointed out that estimating the coordinates from the SIFT descriptors is a very challenging (if possible) problem in \emph{general} settings, as SIFT descriptors could appear anywhere in an image if it is captured in different angles. In other words, for genetic images, the correlation between the SIFT descriptors and the coordinates is actually quite weak. The only hope for relatively accurate estimation of coordinates from SIFT descriptors exists for some highly-structured images, e.g., face images. Specifically, we propose two methods: reference-based and landmark-based approaches for the estimation of coordinates from SIFT descriptors, as demonstrated in Fig. \ref{fig:coordinates_reconstruction}.

\subsubsection{Reference-based Method}

Since it is very challenging to accurately model the relationship between SIFT descriptors and coordinates from one given image, we build up a reference dataset attempting to provide some prior knowledge. Let $\mathcal{F} = \{\mathbf{f}_1, \mathbf{f}_2, \cdots, \mathbf{f}_n\}$ be the given SIFT descriptors. For each SIFT descriptor in $\mathcal{F} $, the straightforward idea is to find the most similar descriptor from the reference dataset using a nearest neighbor (NN) algorithm, and then take its coordinate as the estimated one. Let $\mathcal{R} = \{\hat{\mathbf{I}}_1, \hat{\mathbf{I}}_2, \cdots, \hat{\mathbf{I}}_N\}$ be the reference dataset randomly sampled from the training set. Let also $\hat{\mathcal{F}}_{j} = \{\hat{\mathbf{f}}_1^j, \hat{\mathbf{f}}_2^j, \cdots \}$ be the set of SIFT descriptors extracted from image $\hat{\mathbf{I}}_j$ at coordinates $\{(x_1^j, y_1^j), (x_2^j, y_2^j), \cdots \}$. Define $\hat{\mathcal{F}}$ as the set recording all the SIFT descriptors, namely,


\begin{equation}
\hat{\mathcal{F}} = \bigcup\hat{\mathcal{F}}_j, j=1,\cdots,N.
\end{equation} Then the coordinate of $\mathbf{f}_i \in \mathcal{F}$ can be estimated by using the following NN algorithm:

\begin{equation}\label{eq:coor}
(x^i,y^i) = c(\hat{\mathbf{f}}_j)
\end{equation} where

\begin{equation}\label{eq:mostSimilar}
\hat{\mathbf{f}}_j = \arg\min\limits_{\hat{\mathbf{f}}_j \in \hat{\mathcal{F}}} d(\mathbf{f}_i, \hat{\mathbf{f}}_j)
\end{equation} Here, $d(\cdot)$ computes the Euclidean distance of two descriptors and $c(\cdot)$ returns the coordinate of the input descriptor. Also, the reason why we use the NN algorithm rather than the SIFT matching algorithm is that, in most cases, the input and the reference image do not contain identical objects, making it almost impossible to find a matching pair. In the cases when multiple descriptors are projected to the same coordinate, we randomly keep one descriptor.



This straightforward method can find the globally most similar SIFT descriptors from the reference dataset; however, the recovered features map cannot guarantee the existence of stable object contours, mainly because these coordinates could be obtained from multiple reference images. To mitigate the aforementioned drawback, we propose to use the NN algorithm at the image level instead of the descriptor level. That is, for the whole set of descriptors $\mathcal{F}$, we first find \emph{one} reference $\hat{\mathcal{F}}^*$ with the minimum average distances, and then project each descriptor in $\mathcal{F}$ onto the coordinates of the nearest descriptor within $\hat{\mathcal{F}}^*$. Mathematically,

\begin{equation}
\hat{\mathcal{F}}^* = \arg\min\limits_{\hat{\mathcal{F}_j}} D(\mathcal{F}, \hat{\mathcal{F}_j}), j=1,\cdots,N,
\end{equation}
where
\begin{equation}
D(\mathcal{F}, \hat{\mathcal{F}_j}) = \frac{1}{n}\sum_{i}\min_{\hat{\mathbf{f}}_j}(d(\mathbf{f}_i, \hat{\mathbf{f}}_j)), \mathbf{f}_i \in \mathcal{F}, \hat{\mathbf{f}}_j \in \hat{\mathcal{F}_j}.
\end{equation}
Upon having $\hat{\mathcal{F}}^*$, the coordinate of $\mathbf{f}_i \in \mathcal{F}$ can be similarly estimated as (\ref{eq:coor}) and (\ref{eq:mostSimilar}) by replacing $\hat{\mathcal{F}}$ in (\ref{eq:mostSimilar}) with $\hat{\mathcal{F}}^*$.

We now explain how to form the reference dataset $\mathcal{R}$, which is related to the training set for the deep generative model SLI. In this work, we consider three publicly available datasets \texttt{CelebA-HQ} \cite{karras2017progressive}, \texttt{MD-NYU} \cite{pittaluga2019revealing} and \texttt{ImageNet} \cite{deng2009imagenet}. \texttt{CelebA-HQ} has only one category consisting of face images; \texttt{MD-NYU} has two categories: buildings scenes and indoor scenes; and \texttt{ImageNet} \cite{deng2009imagenet} is much more diverse with one thousand categories. For a given dataset, $\mathcal{R}$ is formed by randomly picking one image from each category. For instance, $\mathcal{R}$ contains only one face image when \texttt{CelebA-HQ} is used, while $\mathcal{R}$ becomes a set with 1000 images in the case of \texttt{ImageNet}. We also have tried to increase the number of images picked from each category, but found that the improvements on the reconstructed images are very slight.

%
%




\subsubsection{Landmark-based Method}

The second method called landmark-based method only works for face images. Specifically, we train a classifier to roughly classify the SIFT descriptors into several pre-defined categories corresponding to different face regions, and then recover the coordinates. The schematic diagram of the landmark-based method is given in Fig. \ref{fig:coordinates_reconstruction} (b). At the training stage, we firstly use Dlib \cite{king2009dlib} to extract landmarks, and then classify the landmarks into seven categories: jaw, right/left brow, nose, right/left eye, mouth (labeled from 0 to 6 respectively). Formally, for a given face image $\mathbf{I}$, Dlib can detect its landmarks $\mathcal{M} = \{(x_i, y_i) | i\in[0, 67]\}$, where each coordinate represents a location of the facial region. For instance, the indexes $\mathbb{R}_0 = [0, 16]$ means jaw region and $\mathbb{R}_1 = [17, 21]$, $\mathbb{R}_2 = [22, 26]$, $\mathbb{R}_3 = [27, 34]$, $\mathbb{R}_4 = [35, 41]$, $\mathbb{R}_5 = [42, 48]$, $\mathbb{R}_6 = [48, 68]$ indicate right/left brow, nose, right/left eye and mouth regions respectively. Then, for each SIFT descriptor, we search the landmark using minimum Euclidean distance and assign the corresponding label to it. In the case that the minimum Euclidean distance is larger than 10, then another label 7 is assigned, which means that this SIFT descriptor belongs to the ``other'' category (i.e., non-facial region).




Upon having the pairs of SIFT descriptor and its label, we train a classifier $C : \mathbb{R}^{1 \times 128} \to \mathbb{R}^{1 \times 8}$ to classify the SIFT descriptors into the aforementioned 8 categories. The classifier $C$ consists of six fully connected layers, where each layer is composed of a linear layer, a Batch Norm \cite{ioofe2015batch} and a ReLU in a sequential manner. For optimization, we adopt the widely used cross entropy loss,
\begin{equation}
\mathcal{L}_e = -\sum_{c=0}^{7}y_c\mbox{log}(C(\mathbf{f})_c),
\end{equation}
where $C(\mathbf{f})_c$ means the probability that the input descriptor $\mathbf{f}$ belongs to the category $c$, and $y_c$ is 1 if the category is the same as the sample category; otherwise 0. In the training process, we randomly select one thousand image (around 120,000 SIFT descriptors) from \texttt{CelebA-HQ}.

Next, by using the landmark $\hat{\mathcal{M}} = \{(\hat{x}_i, \hat{y}_i)\}$ from an image $\hat{\mathbf{I}}$ of the training set as prior knowledge, we can generate coordinates $(\hat{x}_i + \epsilon, \hat{y}_i + \epsilon), i \in \mathbb{R}_c$ for each input SIFT descriptor according to its predicted label $c$, where a randomly generated integer $\epsilon \in [-3, 3]$ is added to reduce collisions. It should be noted that if the predicted label is 7, we simply discard this SIFT descriptor as it does not belong to any specific facial regions.



\subsection{Absence of Descriptors}

We now investigate another scenario of partial SIFT features where the SIFT descriptors are missing while the coordinates of SIFT keypoints are available to the adversary. For instance, SIFT keypoints could be used as robust reference points, in which case the coordinates changes are employed to rectify a distorted image \cite{fang2019screen}. It was also demonstrated that SIFT coordinates can be used for image quality assessment \cite{decombas2012iqa1, kakli2015iqa2}. The scenario with absence of descriptors has been largely neglected by the existing works \cite{weinzaepfel2011reconstructing, dosovitskiy2016inverting, wu2019image, pittaluga2019revealing}, which mainly focused on how the descriptors leak the information of the latent image. If fact, given the set of SIFT coordinates, it is a much less-challenging task to recover the latent image, compared with the case of lacking coordinates. Specifically, we first transform the coordinates into a binary feature map, where 1's are assigned to the locations with SIFT keypoint and 0's elsewhere. This binary feature map can be readily used as input to the second network of our proposed deep generative model SLI, and the first LBP reconstruction network is simply disabled. In addition, the first layer of the generator $G_2$ needs to be modified as one channel $G_2^\prime : \mathbb{R}^{H \times W \times 1} \to \mathbb{R}^{H \times W \times 3}$, so as to fit the dimension of the binary feature map. The other modules of SLI keep unchanged.


\section{Experimental Results}\label{sec:experiments}
The proposed deep generative model SLI is implemented using PyTorch framework. The training is performed on a desktop equipped with a Core-i7 and a single GTX 2080 GPU. The parameters in Adam are $\beta_1=0.5$, $\beta_2=0.999$ and learning rate $r=1\times10^{-4}$. We train the model with the batch size of 1 and the parameters trading off different terms in the loss functions are fixed to be $\lambda_r=100$, $\lambda_{p}=1$, $\lambda_{s}=10$ and $\lambda_a=0.2$. To embrace the concept of reproducible research, the code of our paper is available at: {https://github.com/HighwayWu/SIFT-Reconstruction}.

\begin{table*}[h!]
	\caption{Quantitative comparison of different reconstruction methods over \texttt{CelebA-HQ}, \texttt{MD-NYU} and \texttt{ImageNet} among SIR \cite{desolneux2017stochastic}, IVR \cite{dosovitskiy2016inverting}, INV \cite{pittaluga2019revealing} and our proposed model SLI. $^-$Lower is better. $^+$Higher is better.}
	\centering
	\label{tab:quantitative}
	\begin{tabular}{ccccccccccccc}
		\toprule
		\multirow{2}{*}{Methods} & \multicolumn{4}{c}{\texttt{CelebA-HQ}} & \multicolumn{4}{c}{\texttt{MD-NYU}} & \multicolumn{4}{c}{\texttt{ImageNet}} \\
		
		\cmidrule(r){2-5} \cmidrule(r){6-9} \cmidrule(r){10-13}
		& FID$^-$ & SSIM$^+$ & PSNR$^+$ & PRM$^+$(\%)
		& FID$^-$ & SSIM$^+$ & PSNR$^+$ & PRM$^+$(\%)
		& FID$^-$ & SSIM$^+$ & PSNR$^+$ & PRM$^+$(\%)\\
		\midrule
		SIR \cite{desolneux2017stochastic}
		& 230.5 & 0.547 & 14.12 & 18.13 & 305.5 & 0.271 & 11.18 & 2.08 & 325.0 & 0.325 & 12.77 & 3.18 \\
		IVR \cite{dosovitskiy2016inverting}
		& 143.5 & 0.540 & 17.62 & 25.79 & 363.5 & 0.305 & 13.55 & 1.82 & 294.8 & 0.308 & 14.21 & 8.27 \\
		INV \cite{pittaluga2019revealing}
		& 73.5 & 0.641 & 17.11 & 28.78 & 136.4 & 0.478 & 13.91 & 8.11 & 189.7 & 0.482 & 15.11 & 29.47 \\
		SLI (Ours)
		& \textbf{22.6} & \textbf{0.670} & \textbf{18.95} & \textbf{31.71} & \textbf{119.1} & \textbf{0.485} & \textbf{14.81} & \textbf{10.49} & \textbf{173.4} & \textbf{0.513} & \textbf{15.80} & \textbf{35.92} \\
		\bottomrule
	\end{tabular}
\end{table*}

\begin{figure*}[h!]
	\begin{subfigure}{.162\textwidth}
		\centering
		\includegraphics[width=\textwidth]{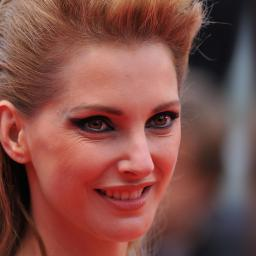}
	\end{subfigure}
	\begin{subfigure}{.162\textwidth}
		\centering
		\includegraphics[width=\textwidth]{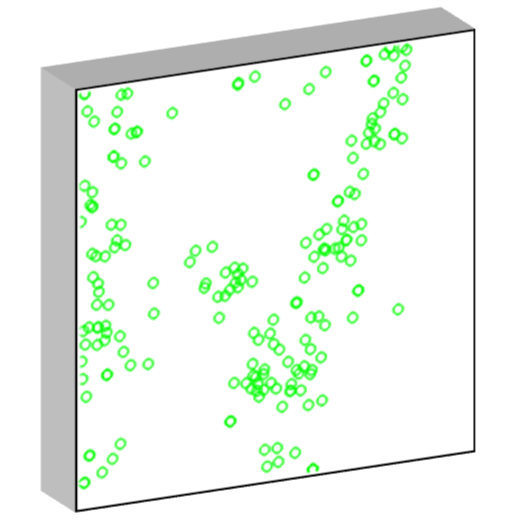}
	\end{subfigure}
	\begin{subfigure}{.162\textwidth}
		\centering
		\includegraphics[width=\textwidth]{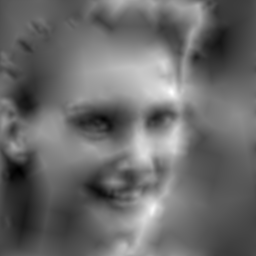}
	\end{subfigure}
	\begin{subfigure}{.162\textwidth}
		\centering
		\includegraphics[width=\textwidth]{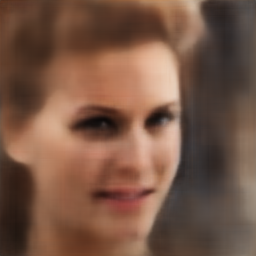}
	\end{subfigure}
	\begin{subfigure}{.162\textwidth}
		\centering
		\includegraphics[width=\textwidth]{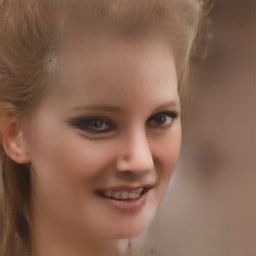}
	\end{subfigure}
	\begin{subfigure}{.162\textwidth}
		\centering
		\includegraphics[width=\textwidth]{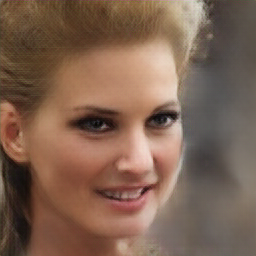}
	\end{subfigure}
	\newline
	\newline
	\begin{subfigure}{.162\textwidth}
		\centering
		\includegraphics[width=\textwidth]{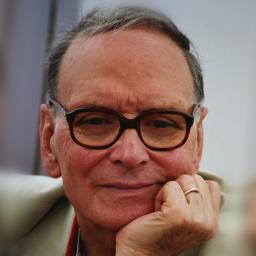}
	\end{subfigure}
	\begin{subfigure}{.162\textwidth}
		\centering
		\includegraphics[width=\textwidth]{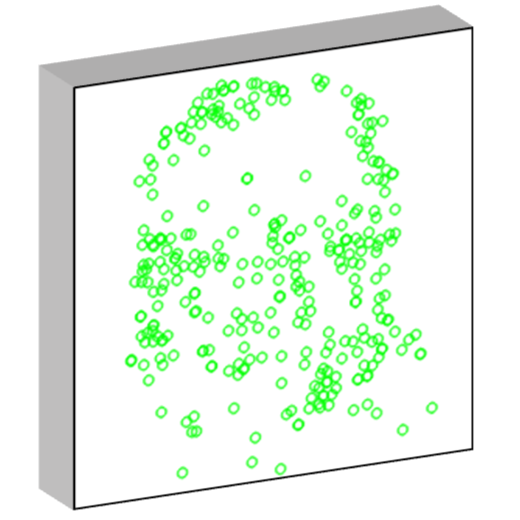}
	\end{subfigure}
	\begin{subfigure}{.162\textwidth}
		\centering
		\includegraphics[width=\textwidth]{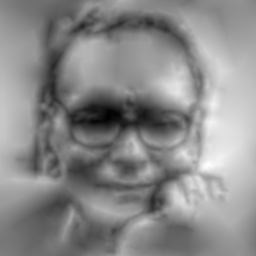}
	\end{subfigure}
	\begin{subfigure}{.162\textwidth}
		\centering
		\includegraphics[width=\textwidth]{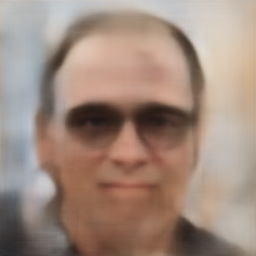}
	\end{subfigure}
	\begin{subfigure}{.162\textwidth}
		\centering
		\includegraphics[width=\textwidth]{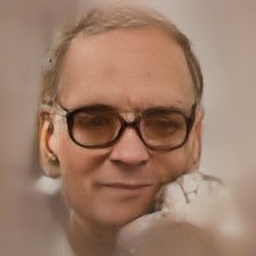}
	\end{subfigure}
	\begin{subfigure}{.162\textwidth}
		\centering
		\includegraphics[width=\textwidth]{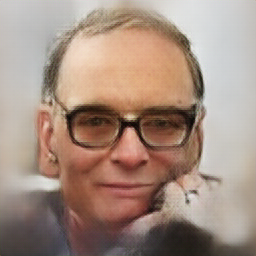}
	\end{subfigure}	
	\newline
	\newline
	\begin{subfigure}{.162\textwidth}
		\centering
		\includegraphics[width=\textwidth]{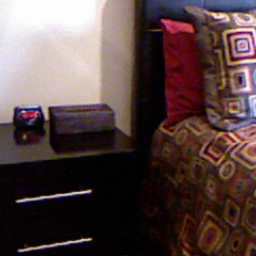}
	\end{subfigure}
	\begin{subfigure}{.162\textwidth}
		\centering
		\includegraphics[width=\textwidth]{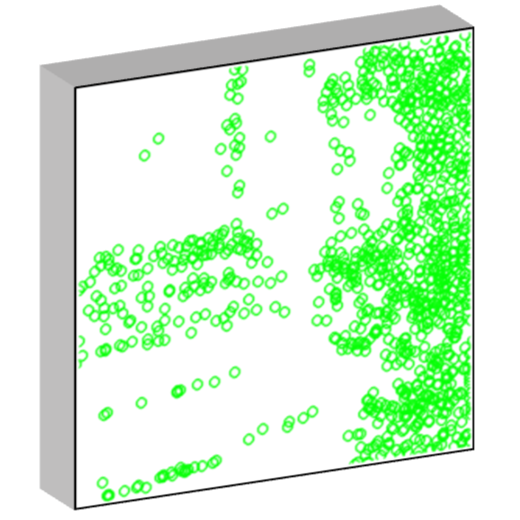}
	\end{subfigure}
	\begin{subfigure}{.162\textwidth}
		\centering
		\includegraphics[width=\textwidth]{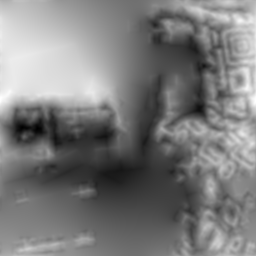}
	\end{subfigure}
	\begin{subfigure}{.162\textwidth}
		\centering
		\includegraphics[width=\textwidth]{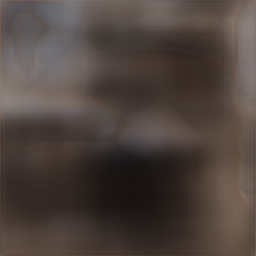}
	\end{subfigure}
	\begin{subfigure}{.162\textwidth}
		\centering
		\includegraphics[width=\textwidth]{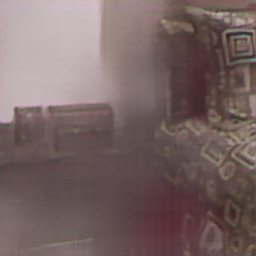}
	\end{subfigure}
	\begin{subfigure}{.162\textwidth}
		\centering
		\includegraphics[width=\textwidth]{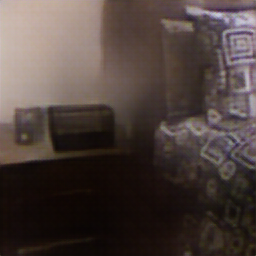}
	\end{subfigure}	
	\newline
	\newline
	\begin{subfigure}{.162\textwidth}
		\centering
		\includegraphics[width=\textwidth]{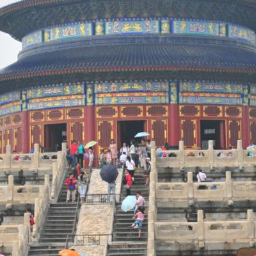}
	\end{subfigure}
	\begin{subfigure}{.162\textwidth}
		\centering
		\includegraphics[width=\textwidth]{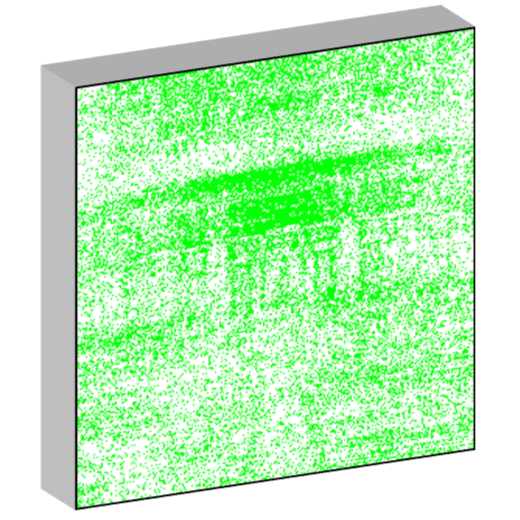}
	\end{subfigure}
	\begin{subfigure}{.162\textwidth}
		\centering
		\includegraphics[width=\textwidth]{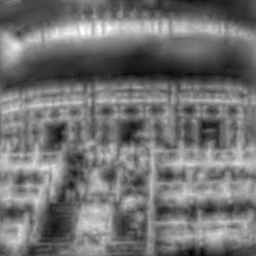}
	\end{subfigure}
	\begin{subfigure}{.162\textwidth}
		\centering
		\includegraphics[width=\textwidth]{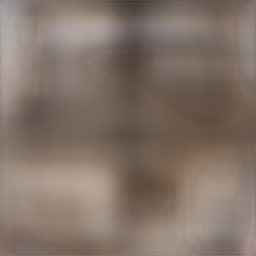}
	\end{subfigure}
	\begin{subfigure}{.162\textwidth}
		\centering
		\includegraphics[width=\textwidth]{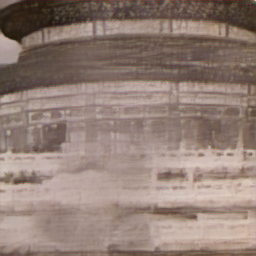}
	\end{subfigure}
	\begin{subfigure}{.162\textwidth}
		\centering
		\includegraphics[width=\textwidth]{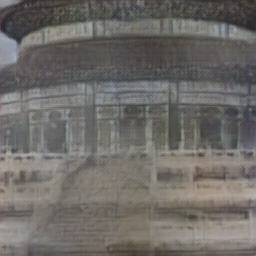}
	\end{subfigure}	
	\newline
	\newline
	\begin{subfigure}{.162\textwidth}
		\centering
		\includegraphics[width=\textwidth]{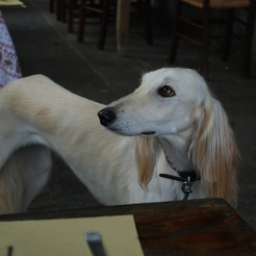}
	\end{subfigure}
	\begin{subfigure}{.162\textwidth}
		\centering
		\includegraphics[width=\textwidth]{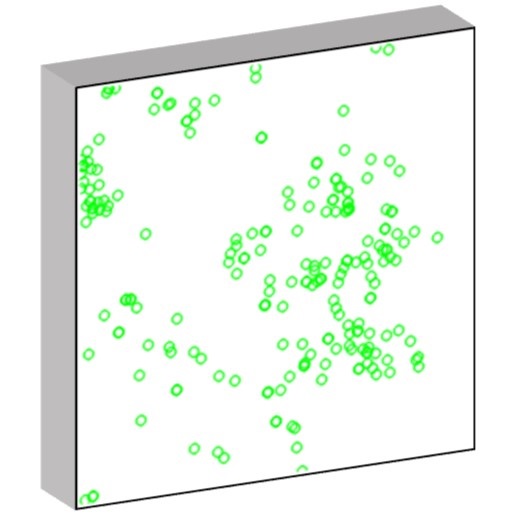}
	\end{subfigure}
	\begin{subfigure}{.162\textwidth}
		\centering
		\includegraphics[width=\textwidth]{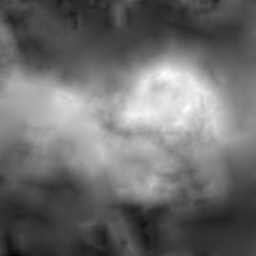}
	\end{subfigure}
	\begin{subfigure}{.162\textwidth}
		\centering
		\includegraphics[width=\textwidth]{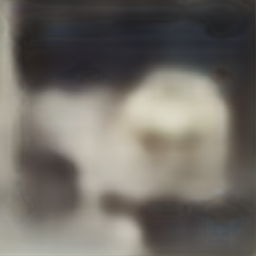}
	\end{subfigure}
	\begin{subfigure}{.162\textwidth}
		\centering
		\includegraphics[width=\textwidth]{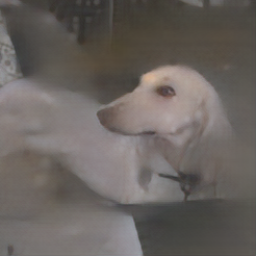}
	\end{subfigure}
	\begin{subfigure}{.162\textwidth}
		\centering
		\includegraphics[width=\textwidth]{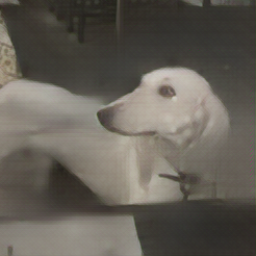}
	\end{subfigure}	
	\newline
	\newline
	\begin{subfigure}{.162\textwidth}
		\centering
		\includegraphics[width=\textwidth]{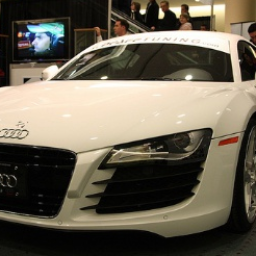}
		\caption{GT.}
	\end{subfigure}
	\begin{subfigure}{.162\textwidth}
		\centering
		\includegraphics[width=\textwidth]{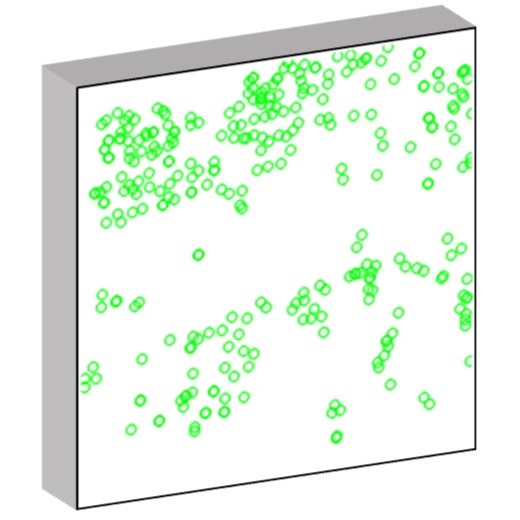}
		\caption{Input}
	\end{subfigure}
	\begin{subfigure}{.162\textwidth}
		\centering
		\includegraphics[width=\textwidth]{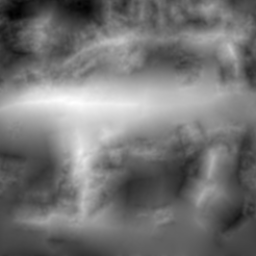}
		\caption{SIR \cite{desolneux2017stochastic}}
	\end{subfigure}
	\begin{subfigure}{.162\textwidth}
		\centering
		\includegraphics[width=\textwidth]{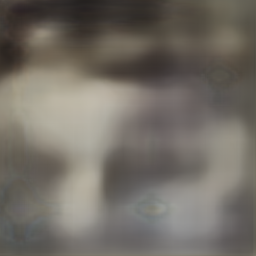}
		\caption{IVR \cite{dosovitskiy2016inverting}}
	\end{subfigure}
	\begin{subfigure}{.162\textwidth}
		\centering
		\includegraphics[width=\textwidth]{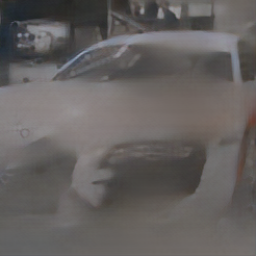}
		\caption{INV \cite{pittaluga2019revealing}}
	\end{subfigure}
	\begin{subfigure}{.162\textwidth}
		\centering
		\includegraphics[width=\textwidth]{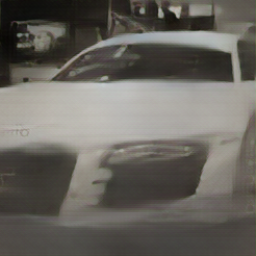}
		\caption{SLI (Ours)}
	\end{subfigure}	
	\caption{Qualitative comparison of different reconstruction methods over \texttt{CelebA-HQ }, \texttt{MD-NYU} and \texttt{ImageNet}. For each row, the images from left to right are ground truth, input SIFT features map, results generated by SIR \cite{desolneux2017stochastic}, IVR \cite{dosovitskiy2016inverting}, INV \cite{pittaluga2019revealing} and the proposed model SLI, respectively.}
	\label{fig:experimental_results}
\end{figure*}

We evaluate the reconstruction performance of our method over three publicly available datasets: a high-quality human face dataset \texttt{CelebA-HQ} \cite{karras2017progressive}, a 3D point cloud dataset containing different indoor and outdoor scenes \texttt{MD-NYU} \cite{pittaluga2019revealing} and a large visualization dataset with one thousand category \texttt{ImageNet} \cite{deng2009imagenet}. The \texttt{CelebA-HQ} dataset contains 28,000 training images and 2000 testing images. The \texttt{MD-NYU} dataset has 8192 images in the training set and 1024 images in the testing set. The \texttt{ImageNet} dataset includes over 1.2 million training images and 100,000 testing images.


\subsection{Evaluations under \textbf{Scenario I}}\label{sec:qualitative}

We first compare the image reconstruction performance of different algorithms under \textbf{Scenario I}. For comparison purpose, we adopt three state-of-the-art SIFT-based image reconstruction methods: Stochastic Image Reconstruction (SIR) \cite{desolneux2017stochastic}, Inverting Visual Representations (IVR) \cite{dosovitskiy2016inverting}, and Revealing Scenes by Inverting (INV)  \cite{pittaluga2019revealing}. Fig. \ref{fig:experimental_results} shows the reconstruction results for some representative testing images. As can be observed, SIR \cite{desolneux2017stochastic} can restore the main semantic information where the SIFT keypoints exist, whereas the areas with insufficient number of SIFT keypoints cannot be recovered satisfactorily. In addition, the reconstructed images lose all the color information. This is because SIR is based on Poisson editing rather than neural networks with training datasets. IVR \cite{dosovitskiy2016inverting} can reconstruct much more realistic color images by using CNNs. However, the reconstructed contents are highly blurry and many fine details are missing. Furthermore, even though INV \cite{pittaluga2019revealing} can produce pretty good results by adopting a deep GAN-based neural network, some broken or blurred textures can be observed. Compared with these methods, our proposed model can learn more reasonable structures and generate more realistic reconstructions, especially those fine structures and texture regions.

%


In addition to the visual comparison of the reconstructed images, we also compare different methods quantitatively, as shown in Table \ref{tab:quantitative}. Here, we adopt the commonly used metrics, namely, structural similarity index (SSIM), peak signal-to-noise ratio (PSNR) and Frechet Inception Distance (FID) \cite{heusel2017fid}. SSIM and PSNR are the most widely used objective measurements of the image quality; however, they may assign inappropriate scores to perceptually accurate results \cite{nazeri2019edgeconnect}. Therefore, FID is often introduced to reflect the Wasserstein-2 distance between the feature space representations of real and generated images using a pre-trained Inception-V3 model \cite{szegedy2016inception}. It can be seen that our method consistently outperforms the competing algorithms under all these three criteria.

\begin{figure*}[t!]
	\begin{subfigure}{\textwidth}
		\begin{subfigure}{.105\textwidth}
			\centering
			\includegraphics[width=\textwidth]{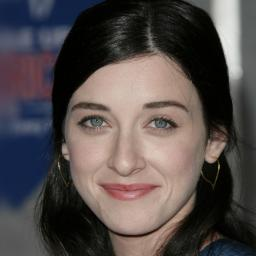}
		\end{subfigure}
		\begin{subfigure}{.105\textwidth}
			\centering
			\includegraphics[width=\textwidth]{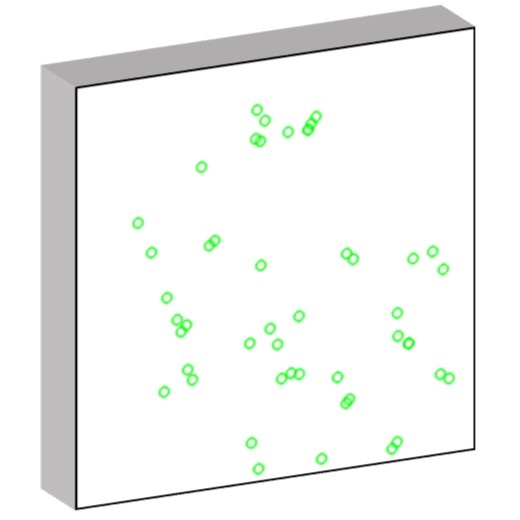}
		\end{subfigure}
		\begin{subfigure}{.105\textwidth}
			\centering
			\includegraphics[width=\textwidth]{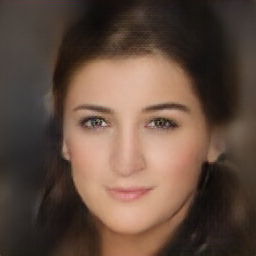}
		\end{subfigure}
		\begin{subfigure}{.105\textwidth}
			\centering
			\includegraphics[width=\textwidth]{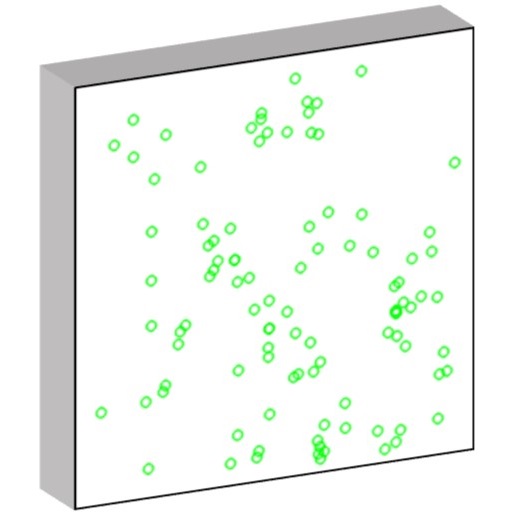}
		\end{subfigure}
		\begin{subfigure}{.105\textwidth}
			\centering
			\includegraphics[width=\textwidth]{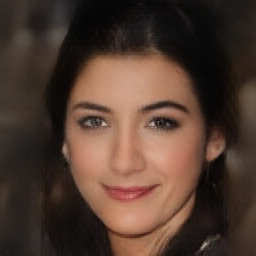}
		\end{subfigure}
		\begin{subfigure}{.105\textwidth}
			\centering
			\includegraphics[width=\textwidth]{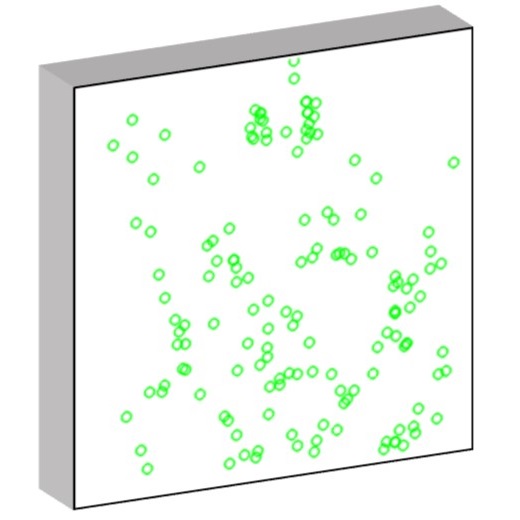}
		\end{subfigure}
		\begin{subfigure}{.105\textwidth}
			\centering
			\includegraphics[width=\textwidth]{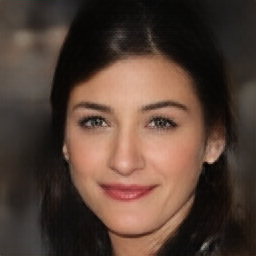}
		\end{subfigure}
		\begin{subfigure}{.105\textwidth}
			\centering
			\includegraphics[width=\textwidth]{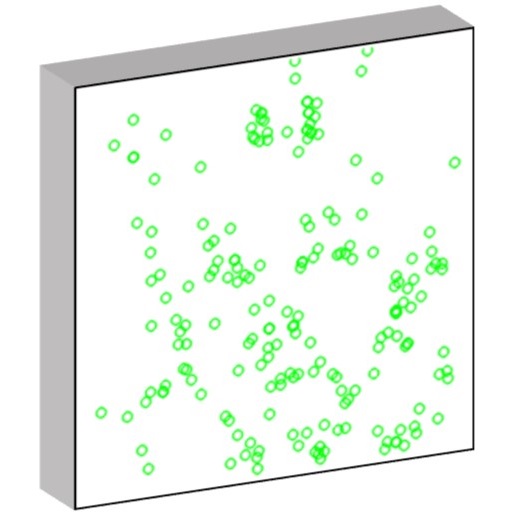}
		\end{subfigure}
		\begin{subfigure}{.105\textwidth}
			\centering
			\includegraphics[width=\textwidth]{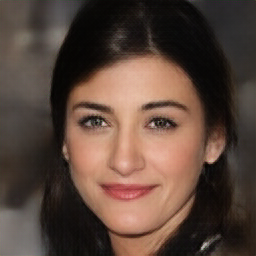}
		\end{subfigure}
	\end{subfigure}
	\newline
	\newline
	\begin{subfigure}{\textwidth}
		\begin{subfigure}{.105\textwidth}
			\centering
			\includegraphics[width=\textwidth]{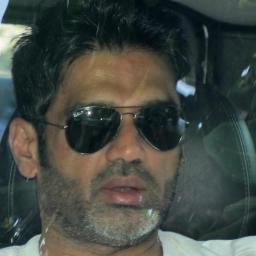}
		\end{subfigure}
		\begin{subfigure}{.105\textwidth}
			\centering
			\includegraphics[width=\textwidth]{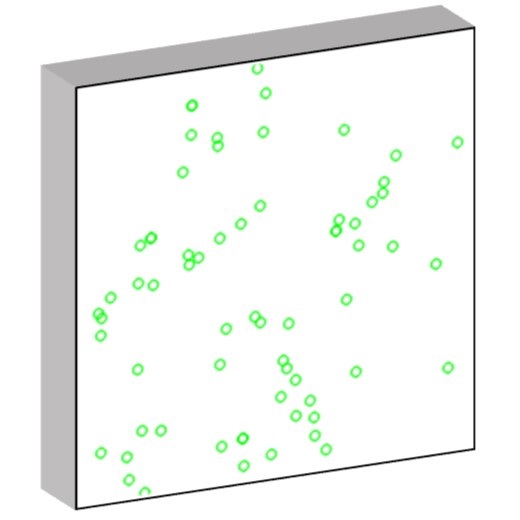}
		\end{subfigure}
		\begin{subfigure}{.105\textwidth}
			\centering
			\includegraphics[width=\textwidth]{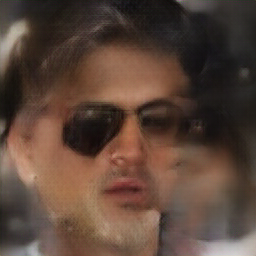}
		\end{subfigure}
		\begin{subfigure}{.105\textwidth}
			\centering
			\includegraphics[width=\textwidth]{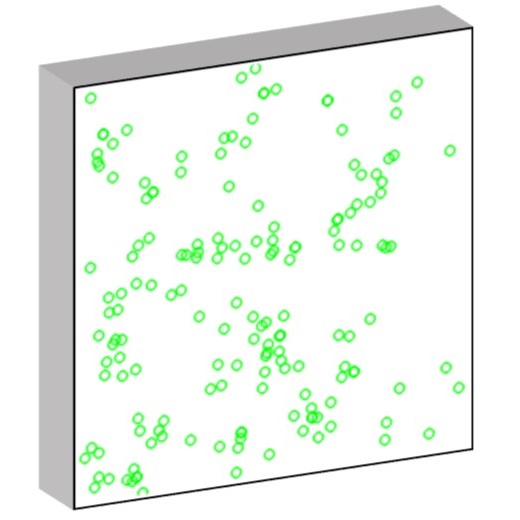}
		\end{subfigure}
		\begin{subfigure}{.105\textwidth}
			\centering
			\includegraphics[width=\textwidth]{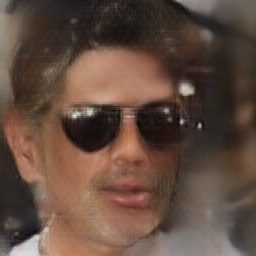}
		\end{subfigure}
		\begin{subfigure}{.105\textwidth}
			\centering
			\includegraphics[width=\textwidth]{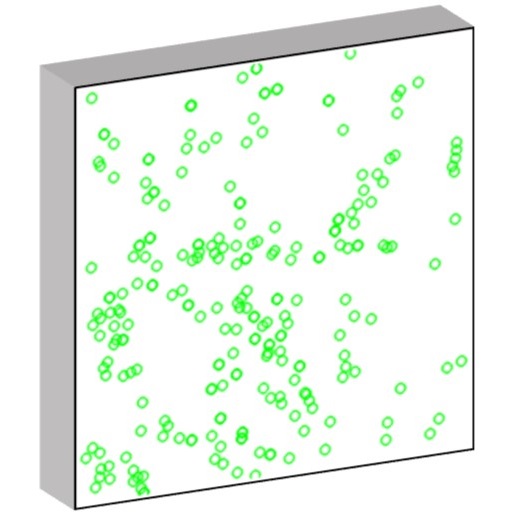}
		\end{subfigure}
		\begin{subfigure}{.105\textwidth}
			\centering
			\includegraphics[width=\textwidth]{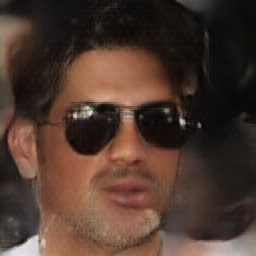}
		\end{subfigure}
		\begin{subfigure}{.105\textwidth}
			\centering
			\includegraphics[width=\textwidth]{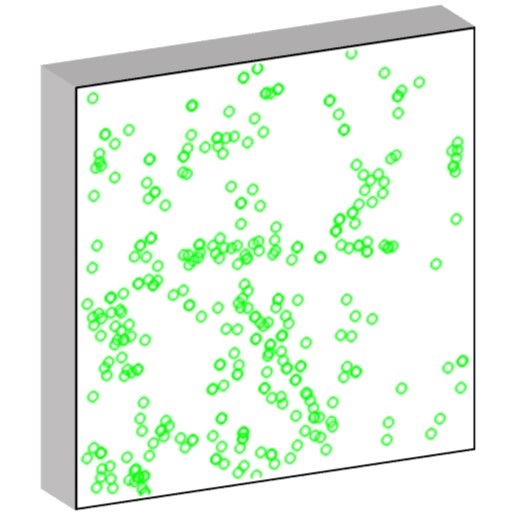}
		\end{subfigure}
		\begin{subfigure}{.105\textwidth}
			\centering
			\includegraphics[width=\textwidth]{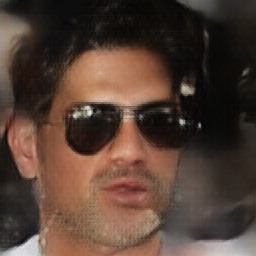}
		\end{subfigure}
	\end{subfigure}
	\newline
	\newline
	\begin{subfigure}{\textwidth}
		\begin{subfigure}{.105\textwidth}
			\centering
			\includegraphics[width=\textwidth]{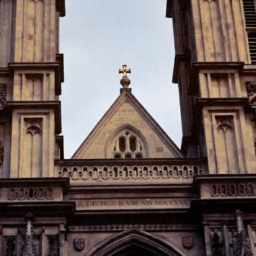}
			\caption{GT.}
		\end{subfigure}
		\begin{subfigure}{.105\textwidth}
			\centering
			\includegraphics[width=\textwidth]{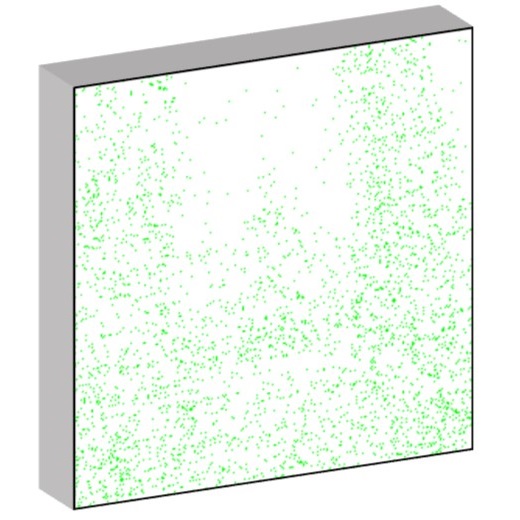}
			\caption{In. (25\%)}
		\end{subfigure}
		\begin{subfigure}{.105\textwidth}
			\centering
			\includegraphics[width=\textwidth]{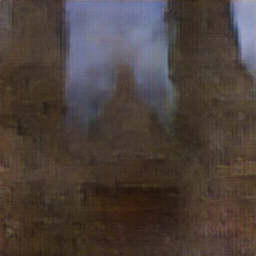}
			\caption{Res. (25\%)}
		\end{subfigure}
		\begin{subfigure}{.105\textwidth}
			\centering
			\includegraphics[width=\textwidth]{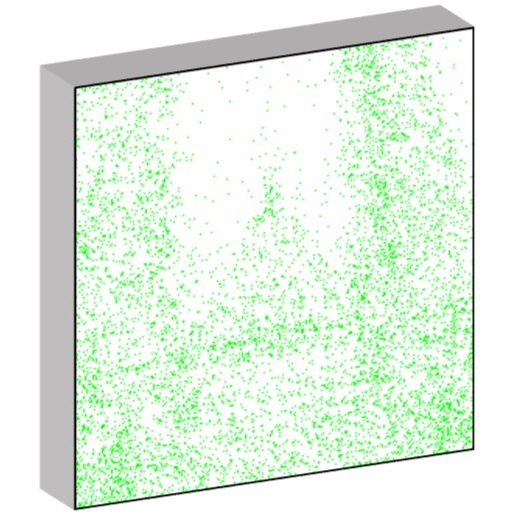}
			\caption{In. (50\%)}
		\end{subfigure}
		\begin{subfigure}{.105\textwidth}
			\centering
			\includegraphics[width=\textwidth]{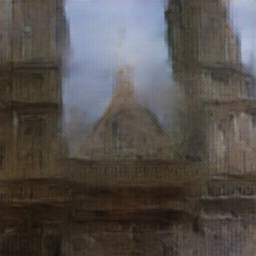}
			\caption{Res. (50\%)}
		\end{subfigure}
		\begin{subfigure}{.105\textwidth}
			\centering
			\includegraphics[width=\textwidth]{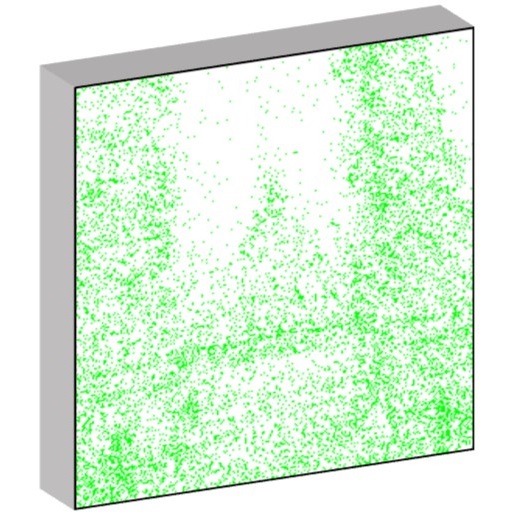}
			\caption{In. (75\%)}
		\end{subfigure}
		\begin{subfigure}{.105\textwidth}
			\centering
			\includegraphics[width=\textwidth]{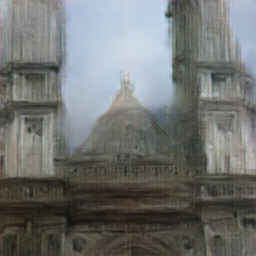}
			\caption{Res. (75\%)}
		\end{subfigure}
		\begin{subfigure}{.105\textwidth}
			\centering
			\includegraphics[width=\textwidth]{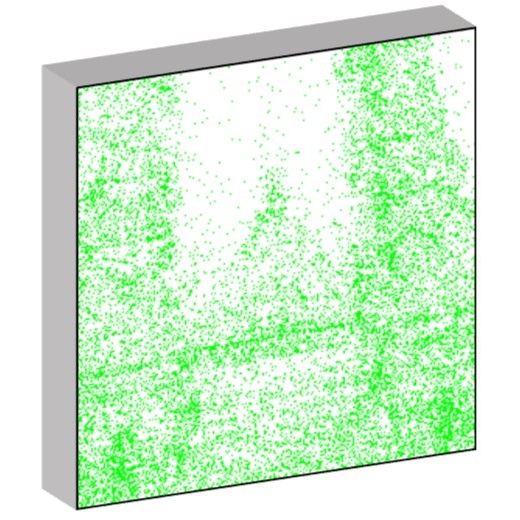}
			\caption{In. (100\%)}
		\end{subfigure}
		\begin{subfigure}{.105\textwidth}
			\centering
			\includegraphics[width=\textwidth]{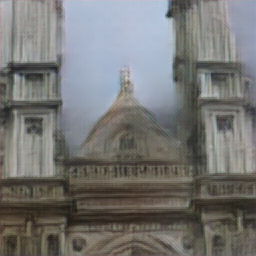}
			\caption{Res.(100\%)}
		\end{subfigure}
	\end{subfigure}
	\renewcommand\thefigure{7}
	\caption{Robustness evaluation by using different percentage of SIFT features as input. (b), (d), (f), (h) are the input SIFT features map with 25\%, 50\%, 75\% and 100\% of the original SIFT features, and (c), (e), (g), (i) are corresponding reconstruction results.}
	\label{fig:robustness}
\end{figure*}

\begin{figure}[t!]
	\centering
	\begin{subfigure}{0.24\textwidth}
		\centering
		\includegraphics[width=\textwidth]{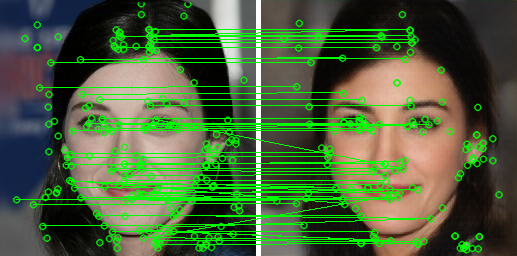}
	\end{subfigure}
	\begin{subfigure}{0.24\textwidth}
		\centering
		\includegraphics[width=\textwidth]{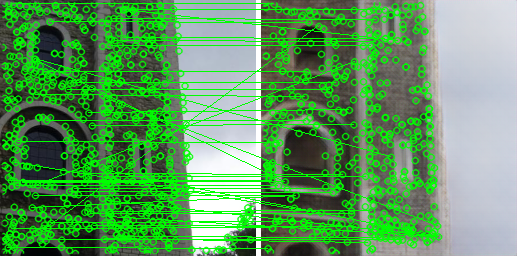}
	\end{subfigure}
	\renewcommand\thefigure{6}
	\caption{Re-matching examples. In each pair, left is the ground truth and right is the reconstruction result. Green lines represent matched SIFT pairs.}
	\label{fig:re_matching}
\end{figure}

Beyond the above traditional metrics for quantitative comparisons, we propose an additional metric by evaluating the percentage of re-matching (for short, PRM) between the ground truth SIFT descriptors and ones from the reconstructed results. This reflects how the reconstructed image preserves the fidelity of the latent image in the SIFT descriptor domain. More specifically, define  $\mathcal{F}_g=\{\mathbf{f}_1^g, \mathbf{f}_2^g, \cdots, \mathbf{f}_m^g\}$ as the set of ground truth descriptors, and $\mathcal{F}_o=\{\mathbf{f}_1^o, \mathbf{f}_2^o, \cdots, \mathbf{f}_n^o\}$ as the set of reconstructed ones. Let $d_{i, 1}$ and $d_{i, 2}$ record the nearest and second-nearest Euclidean distances between the reconstructed descriptor $\mathbf{f}_i^o$ ($i \in [1, n]$) and the ground truth descriptors $\{\mathbf{f}_j^g \vert j \in [1, m]\}$. Then the PRM is defined as:

\begin{equation}
\mathrm{PRM}=\frac{1}{n}\sum_{i=1}^{n}T(d_{i,1}/d_{i,2}, t),
\end{equation}
where $T$ is a thresholding function incorporating the SIFT matching algorithm \cite{lowe2004distinctive},
\begin{equation}
T(d_{i,1}/d_{i,2}, t) =
\begin{cases}
1& \mathrm{if}~ d_{i,1}/d_{i,2} < t\\
0& \mathrm{if}~ d_{i,1}/d_{i,2} \ge t
\end{cases}.
\end{equation}
Here, $t$ is set to 0.8 for guaranteeing the reliable matching according to \cite{lowe2004distinctive}. Obviously, $\mathrm{PRM}$ takes a value in $[0, 1]$, representing the fidelity of the SIFT descriptors extracted from the reconstructed image. The $\mathrm{PRM}$ results of different methods are also compiled into Table \ref{tab:quantitative}. As can be seen, the proposed SLI achieves the best $\mathrm{PRM}$ performance among all the competing algorithms over three test datasets. Re-matching examples are also given in Fig. \ref{fig:re_matching}, where the green lines represent the matched SIFT pairs and the remaining isolated points indicate no match.

\begin{figure}[t]
	\centering
	\begin{subfigure}{0.5\textwidth}
		\begin{subfigure}{.19\textwidth}
			\centering
			\includegraphics[width=\textwidth]{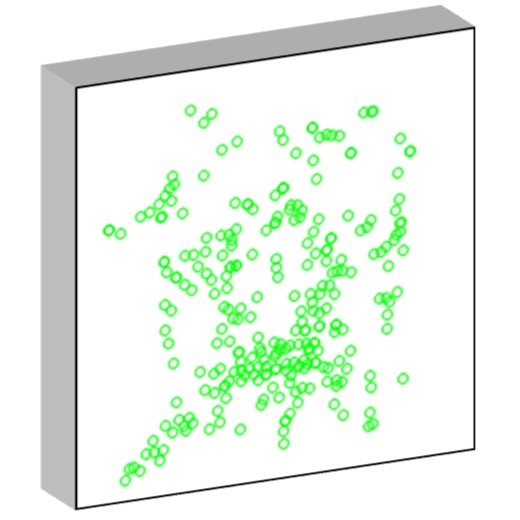}
			\includegraphics[width=\textwidth]{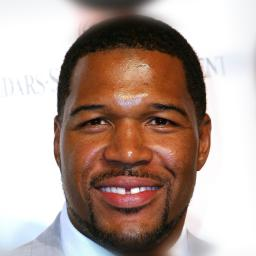}
		\end{subfigure}
		\begin{subfigure}{.19\textwidth}
			\centering
			\includegraphics[width=\textwidth]{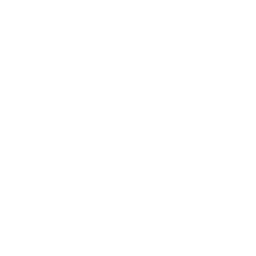}
			\includegraphics[width=\textwidth]{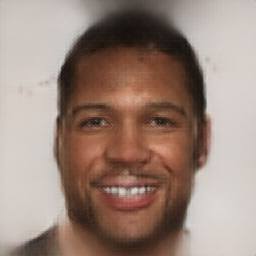}
		\end{subfigure}
		\begin{subfigure}{.19\textwidth}
			\centering
			\includegraphics[width=\textwidth]{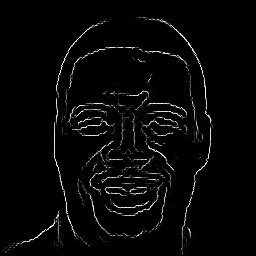}
			\includegraphics[width=\textwidth]{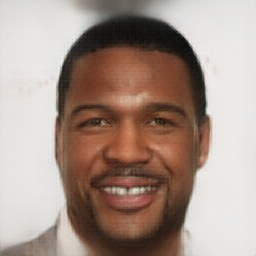}
		\end{subfigure}
		\begin{subfigure}{.19\textwidth}
			\centering
			\includegraphics[width=\textwidth]{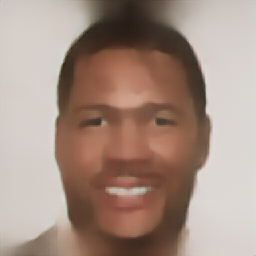}
			\includegraphics[width=\textwidth]{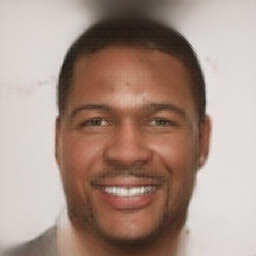}
		\end{subfigure}
		\begin{subfigure}{.19\textwidth}
			\centering
			\includegraphics[width=\textwidth]{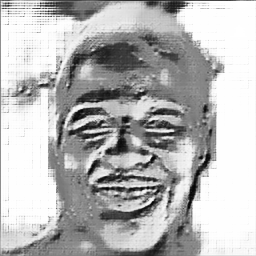}
			\includegraphics[width=\textwidth]{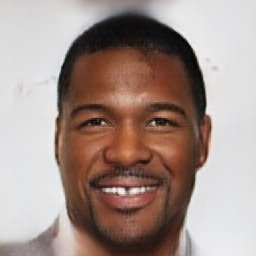}
		\end{subfigure}
	\end{subfigure}
	\begin{subfigure}{0.5\textwidth}
		\begin{subfigure}{.19\textwidth}
			\centering
			\includegraphics[width=\textwidth]{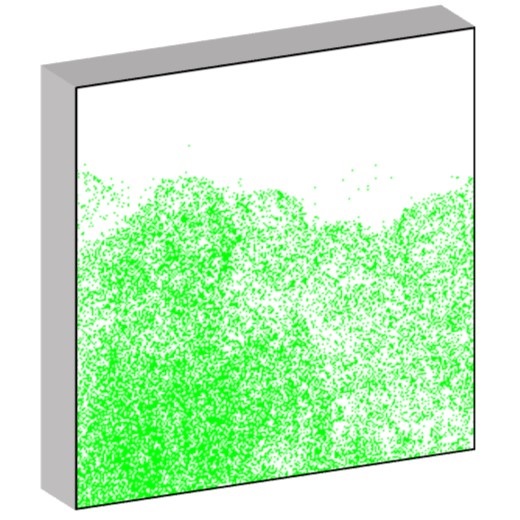}
			\includegraphics[width=\textwidth]{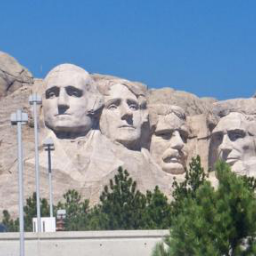}
			\caption{GT.}
		\end{subfigure}
		\begin{subfigure}{.19\textwidth}
			\centering
			\includegraphics[width=\textwidth]{imgs/ablation/empty.png}
			\includegraphics[width=\textwidth]{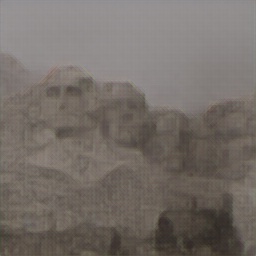}
			\caption{wo}
		\end{subfigure}
		\begin{subfigure}{.19\textwidth}
			\centering
			\includegraphics[width=\textwidth]{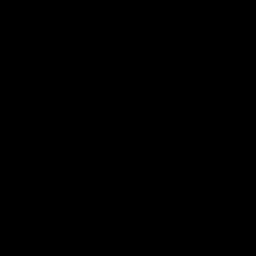}
			\includegraphics[width=\textwidth]{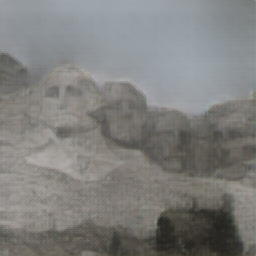}
			\caption{w Edge}
		\end{subfigure}
		\begin{subfigure}{.19\textwidth}
			\centering
			\includegraphics[width=\textwidth]{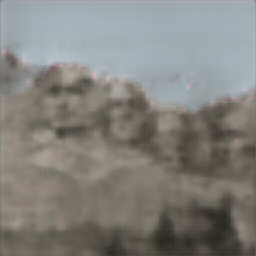}
			\includegraphics[width=\textwidth]{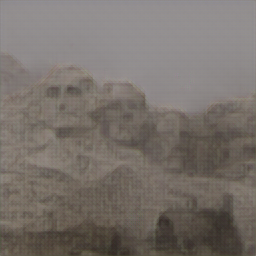}
			\caption{w RTV}
		\end{subfigure}
		\begin{subfigure}{.19\textwidth}
			\centering
			\includegraphics[width=\textwidth]{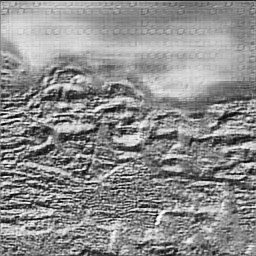}
			\includegraphics[width=\textwidth]{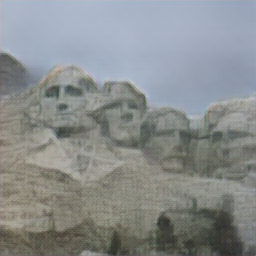}
			\caption{w LBP}
		\end{subfigure}
	\end{subfigure}
	\caption{Effect of the guidance provided by different structures. (a) Input and ground truth. (b)-(e) The first and third rows show the inputs with empty, predicted Canny edges \cite{canny1986computational}, RTV \cite{xu2012rtv} and LBP \cite{ojala1996comparative}, respectively; the second and fourth rows present the corresponding reconstructed results.}
	\label{fig:ablation}
\end{figure}

%


Before ending the discussions under \textbf{Scenario I}, we evaluate the robustness of our proposed SLI. Fig. \ref{fig:robustness} shows the results of randomly using 25\%, 50\%, 75\% and 100\% SIFT features as the model input. Although the input has very high spatial sparsity (e.g., 25\% or 50\% features), the output images are still quite interpretable. These results indicate that the privacy leakage problem is very severe under \textbf{Scenario I}, as only a small portion of the SIFT features could lead to the disclosure of sensitive information.


\subsection{Ablation Studies of SLI}\label{sec:ablation}

We now conduct the ablation studies of our proposed SLI by analyzing how the LBP reconstruction network contributes to the final reconstruction. To this end, we retrain the model without the assistance of the LBP reconstruction network. Further, we consider to replace the LBP reconstruction network with some alternatives including the Canny edges \cite{canny1986computational} and RTV \cite{xu2012rtv}, which could also offer structural information for the image reconstruction \cite{nazeri2019edgeconnect, ren2019structureflow}.

\begin{table}[t]
	\caption{Quantitative comparisons of the guidance provided by Canny edge \cite{canny1986computational}, RTV \cite{xu2012rtv} and LBP \cite{ojala1996comparative} respectively. $^-$Lower is better. $^+$Higher is better.}
	\centering
	\label{tab:ablation}
	\begin{tabular}{ccccc}
		\toprule
		\multirow{2}{*}{Methods} & \multicolumn{4}{c}{\texttt{CelebA-HQ}} \\
		
		\cmidrule(r){2-5}
		& FID$^-$ & SSIM$^+$ & PSNR$^+$ & PRM$^+$(\%)\\
		\midrule
		wo & 42.5 & 0.591 & 17.58 & 20.84 \\
		w Edge & 27.9  & 0.634 & 18.55 & 29.26 \\
		w RTV & 33.8 & 0.605 & 17.89 & 26.19 \\
		w LBP & \textbf{22.6} & \textbf{0.670} & \textbf{18.95} & \textbf{31.71} \\
		\midrule
		\midrule
		\multirow{2}{*}{Methods} & \multicolumn{4}{c}{\texttt{MD-NYU}} \\
		\cmidrule(r){2-5}
		& FID$^-$ & SSIM$^+$ & PSNR$^+$ & PRM$^+$(\%)\\
		\midrule
		wo & 213.3 & 0.398 & 13.49 & 5.34 \\
		w Edge & 149.0 & 0.451 & 14.44 & 6.82 \\
		w RTV & 200.7 & 0.408 & 13.89 & 5.84 \\
		w LBP & \textbf{119.1} & \textbf{0.485} & \textbf{14.81} & \textbf{10.49}\\		
		\bottomrule
	\end{tabular}
\end{table}

The reconstruction results produced with different structural information are demonstrated in Fig. \ref{fig:ablation}. In many cases, the SIFT keypoints are poorly localized along an edge \cite{lowe2004distinctive}, or are too dense to be separated from the edge, making the transform from SIFT to edges inaccurate (e.g., the transformed edges in the first and third rows). Although RTV is a good representation of the global structures, the high-frequency information discarded by RTV results in unsatisfactory outputs.  Meanwhile, from the perspective of the practical implementation, Canny edges and RTV extractions typically involve many parameters (e.g., the pre-filtering strength, the threshold for Canny edges, and the degree of smooth/sharpness for RTV), whose optimal setting should vary for different images. In contrast, LBP is easy to be computed and could be parameter-free. Also, the sufficient information (e.g., gradients) contained in LBP guides the learning direction better and makes the result sharper (e.g., eyes and nose), which could be further validated by the statistical reports in Table \ref{tab:ablation}. These observations would suggest that LBP is a more appropriate candidate for providing structural information in the case of image reconstruction from SIFT features.

\subsection{Evaluations under \textbf{Scenario II}}\label{sec:recon_des}

\begin{figure}[t!]
	\centering
	\begin{subfigure}{0.5\textwidth}
		\centering
		\begin{subfigure}{.13\textwidth}
			\centering
			\includegraphics[width=\textwidth]{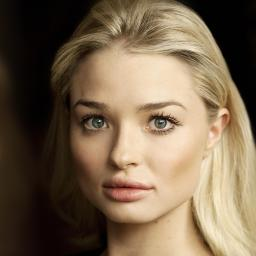}
		\end{subfigure}
		\begin{subfigure}{.13\textwidth}
			\centering
			\includegraphics[width=\textwidth]{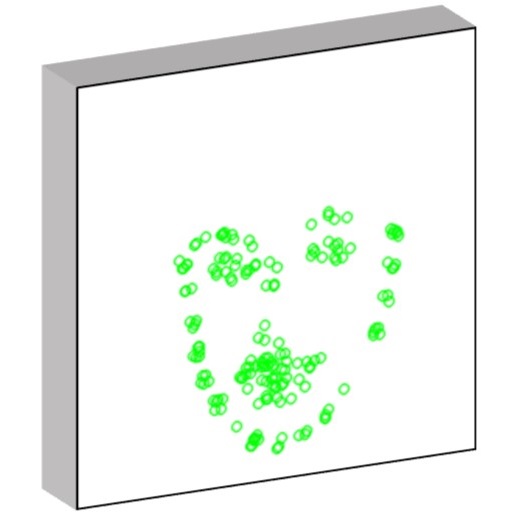}
		\end{subfigure}	
		\begin{subfigure}{.13\textwidth}
			\centering
			\includegraphics[width=\textwidth]{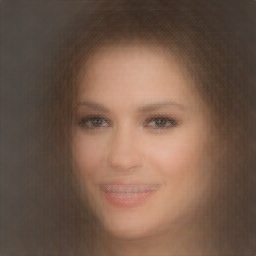}
		\end{subfigure}
		\begin{subfigure}{.13\textwidth}
			\centering
			\includegraphics[width=\textwidth]{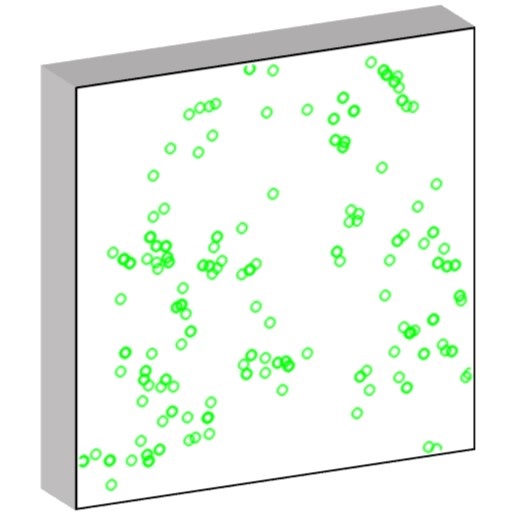}
		\end{subfigure}
		\begin{subfigure}{.13\textwidth}
			\centering
			\includegraphics[width=\textwidth]{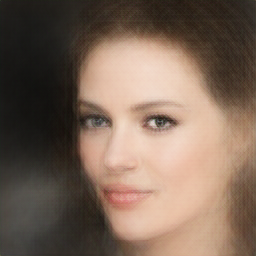}
		\end{subfigure}
		\begin{subfigure}{.13\textwidth}
			\centering
			\includegraphics[width=\textwidth]{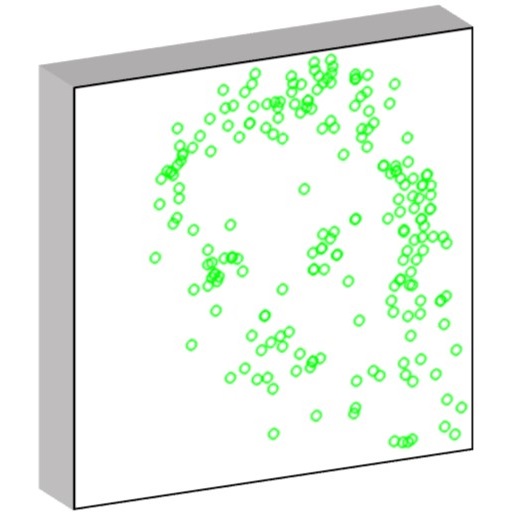}
		\end{subfigure}
		\begin{subfigure}{.13\textwidth}
			\centering
			\includegraphics[width=\textwidth]{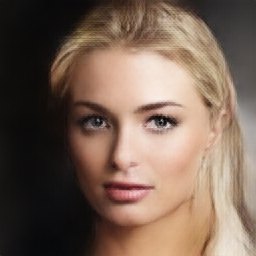}
		\end{subfigure}
	\end{subfigure}
	\begin{subfigure}{0.5\textwidth}
		\centering
		\begin{subfigure}{.13\textwidth}
			\centering
			\includegraphics[width=\textwidth]{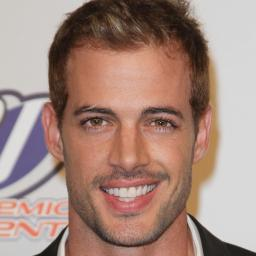}
		\end{subfigure}
		\begin{subfigure}{.13\textwidth}
			\centering
			\includegraphics[width=\textwidth]{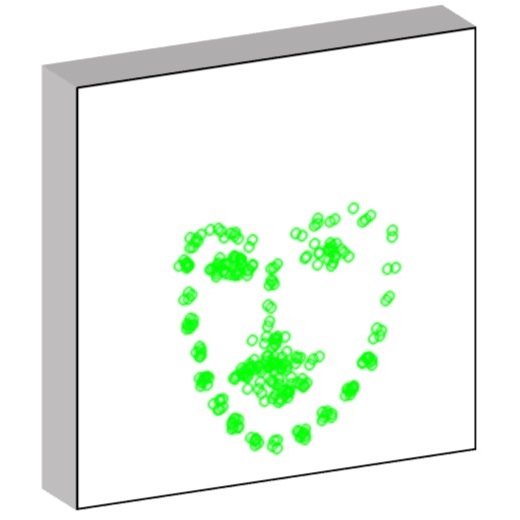}
		\end{subfigure}	
		\begin{subfigure}{.13\textwidth}
			\centering
			\includegraphics[width=\textwidth]{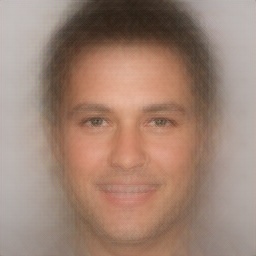}
		\end{subfigure}
		\begin{subfigure}{.13\textwidth}
			\centering
			\includegraphics[width=\textwidth]{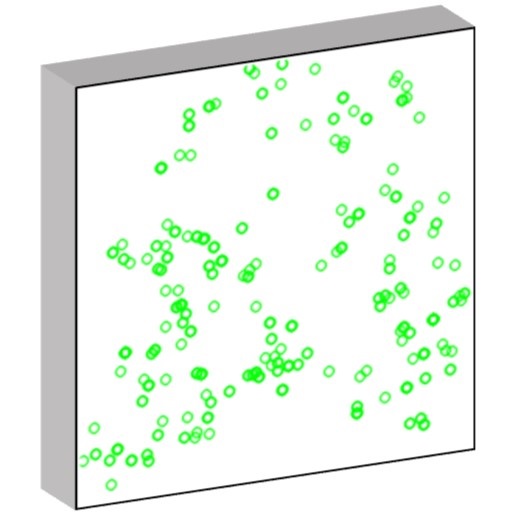}
		\end{subfigure}
		\begin{subfigure}{.13\textwidth}
			\centering
			\includegraphics[width=\textwidth]{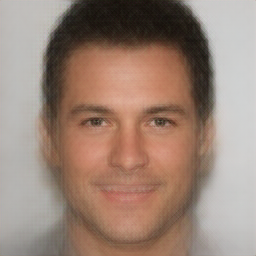}
		\end{subfigure}
		\begin{subfigure}{.13\textwidth}
			\centering
			\includegraphics[width=\textwidth]{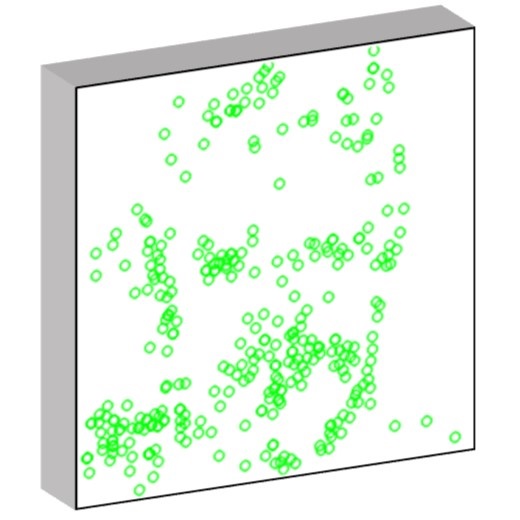}
		\end{subfigure}
		\begin{subfigure}{.13\textwidth}
			\centering
			\includegraphics[width=\textwidth]{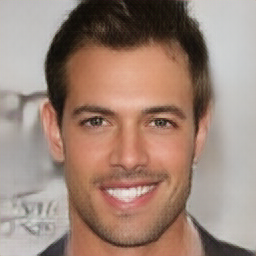}
		\end{subfigure}
	\end{subfigure}
	\begin{subfigure}{0.5\textwidth}
		\centering
		\begin{subfigure}{.13\textwidth}
			\centering
			\includegraphics[width=\textwidth]{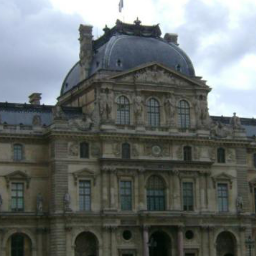}
			\caption{}
		\end{subfigure}
		\begin{subfigure}{.13\textwidth}
			\centering
			\includegraphics[width=\textwidth]{imgs/ablation/empty.png}
			\caption{}
		\end{subfigure}	
		\begin{subfigure}{.13\textwidth}
			\centering
			\includegraphics[width=\textwidth]{imgs/ablation/empty.png}
			\caption{}
		\end{subfigure}
		\begin{subfigure}{.13\textwidth}
			\centering
			\includegraphics[width=\textwidth]{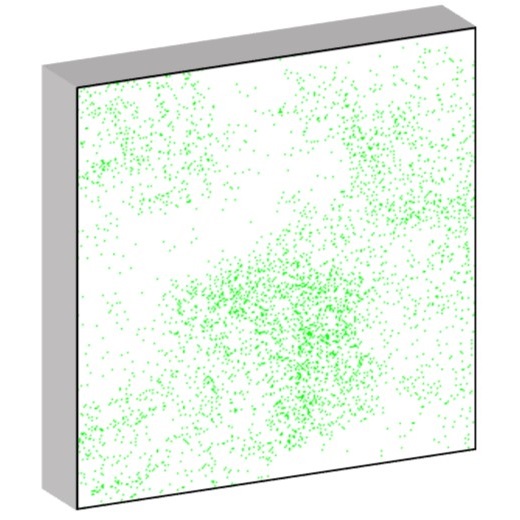}
			\caption{}
		\end{subfigure}
		\begin{subfigure}{.13\textwidth}
			\centering
			\includegraphics[width=\textwidth]{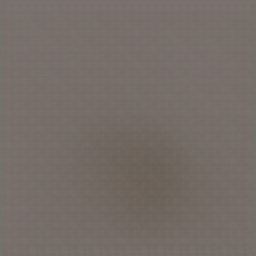}
			\caption{}
		\end{subfigure}
		\begin{subfigure}{.13\textwidth}
			\centering
			\includegraphics[width=\textwidth]{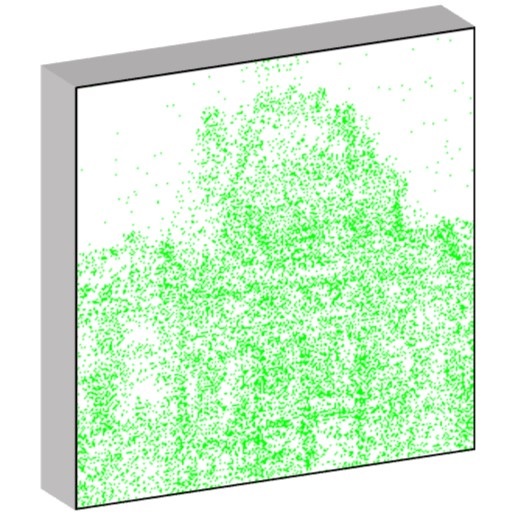}
			\caption{}
		\end{subfigure}
		\begin{subfigure}{.13\textwidth}
			\centering
			\includegraphics[width=\textwidth]{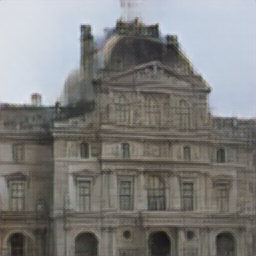}
			\caption{}
		\end{subfigure}
	\end{subfigure}
	\caption{Image reconstruction from solely SIFT descriptors. (a) Ground truth. (b)-(c) Inputs and results of SLI-L. (d)-(e) Inputs and results of SLI-R. (f)-(g) Results of SLI with full SIFT features for comparison.}
	\label{fig:recon_des}
\end{figure}

We now evaluate the performance of the image reconstruction from SIFT features under \textbf{Scenario II}, i.e., either absence of coordinates or absence of descriptors. We first try to reconstruct the image by using \textit{solely} SIFT descriptors as input, in which case the coordinates can be estimated through the reference-based and landmark-based methods presented in Section IV. The reconstruction results are illustrated in Fig. \ref{fig:recon_des}. For simplicity, we call the SLI model with coordinates estimated by the reference-based and landmark-based methods SLI-R and SLI-L, respectively. As can be observed, SLI-L can restore the main semantic information of the facial area, but the results are quite blurry. In contrast, SLI-R can generate sharper and more realistic reconstruction results, primarily thanks to the employment of a reference image. However, both SLI-L and SLI-R have a fatal problem, i.e., choosing a suitable landmark or reference is a crucial issue. As the data included in \texttt{CelebA-HQ} usually have the same skeleton (e.g., eyes, nose and mouth), we can easily project the input descriptors to the corresponding positions in the landmark or reference, while for the dataset (e.g., \texttt{MD-NYU} or \texttt{ImageNet}) that usually contains various categories, it is difficult to find one or more suitable images as the skeleton of SIFT descriptors. As also mentioned in Section \ref{sec:advanced_methods}, SLI-L and SLI-R could fail for generic images without regular structures. The last row of Fig. \ref{fig:recon_des} shows an example of such failure. Besides, the quantitative comparison of SLI-L and SLI-R are reported in Table \ref{tab:advanced}. It is found that SLI-R performs much better than SLI-L with 0.9 dB PSNR gain over \texttt{CelebA-HQ}. Also, as expected, they both perform poorly in the other datasets.

\begin{figure}[t!]
	\centering
	\begin{subfigure}{0.48\textwidth}
		\begin{subfigure}{.19\textwidth}
			\centering
			\includegraphics[width=\textwidth]{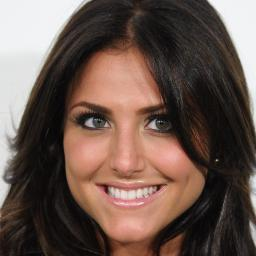}
		\end{subfigure}
		\begin{subfigure}{.19\textwidth}
			\centering
			\includegraphics[width=\textwidth]{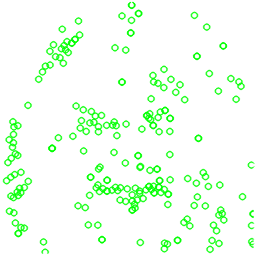}
		\end{subfigure}
		\begin{subfigure}{.19\textwidth}
			\centering
			\includegraphics[width=\textwidth]{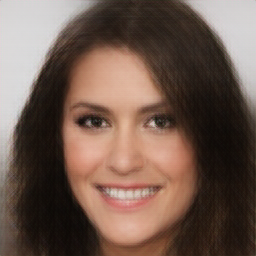}
		\end{subfigure}
		\begin{subfigure}{.19\textwidth}
			\centering
			\includegraphics[width=\textwidth]{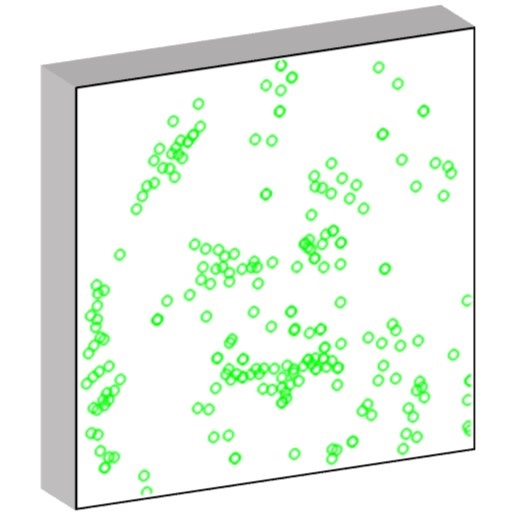}
		\end{subfigure}
		\begin{subfigure}{.19\textwidth}
			\centering
			\includegraphics[width=\textwidth]{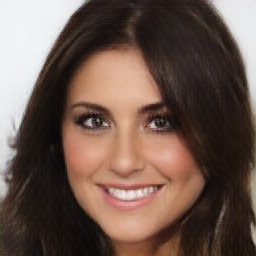}
		\end{subfigure}
	\end{subfigure}
	\begin{subfigure}{0.48\textwidth}
		\begin{subfigure}{.19\textwidth}
			\centering
			\includegraphics[width=\textwidth]{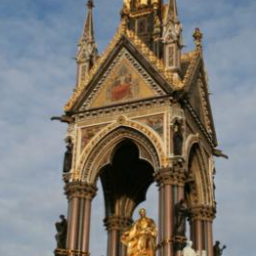}
		\end{subfigure}
		\begin{subfigure}{.19\textwidth}
			\centering
			\includegraphics[width=\textwidth]{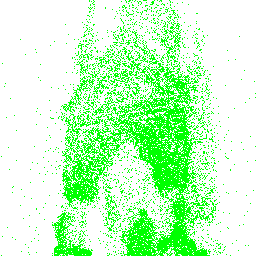}
		\end{subfigure}
		\begin{subfigure}{.19\textwidth}
			\centering
			\includegraphics[width=\textwidth]{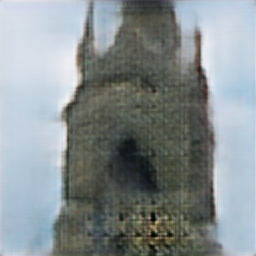}
		\end{subfigure}
		\begin{subfigure}{.19\textwidth}
			\centering
			\includegraphics[width=\textwidth]{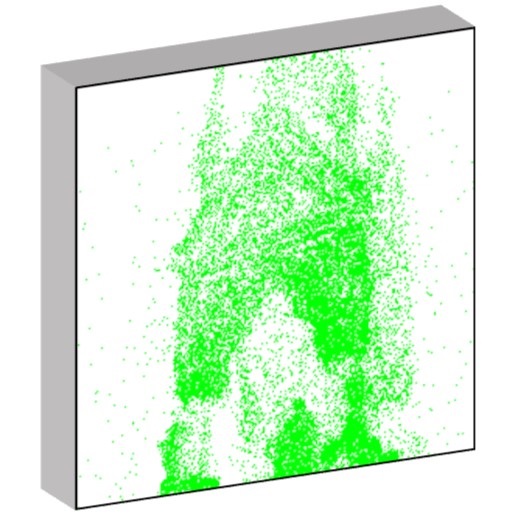}
		\end{subfigure}
		\begin{subfigure}{.19\textwidth}
			\centering
			\includegraphics[width=\textwidth]{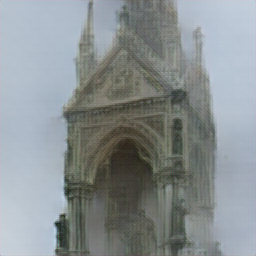}
		\end{subfigure}
	\end{subfigure}
	\begin{subfigure}{0.48\textwidth}
		\begin{subfigure}{.19\textwidth}
			\centering
			\includegraphics[width=\textwidth]{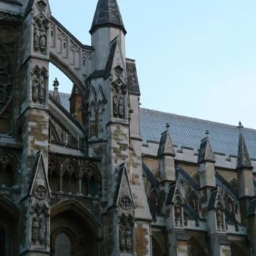}
			\caption{}
		\end{subfigure}
		\begin{subfigure}{.19\textwidth}
			\centering
			\includegraphics[width=\textwidth]{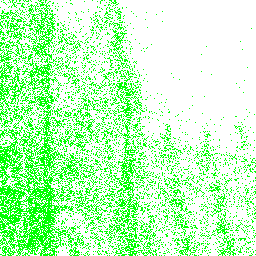}
			\caption{}
		\end{subfigure}
		\begin{subfigure}{.19\textwidth}
			\centering
			\includegraphics[width=\textwidth]{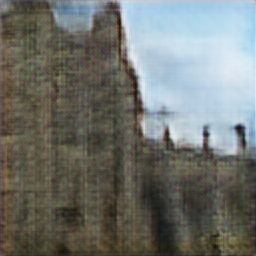}
			\caption{}
		\end{subfigure}
		\begin{subfigure}{.19\textwidth}
			\centering
			\includegraphics[width=\textwidth]{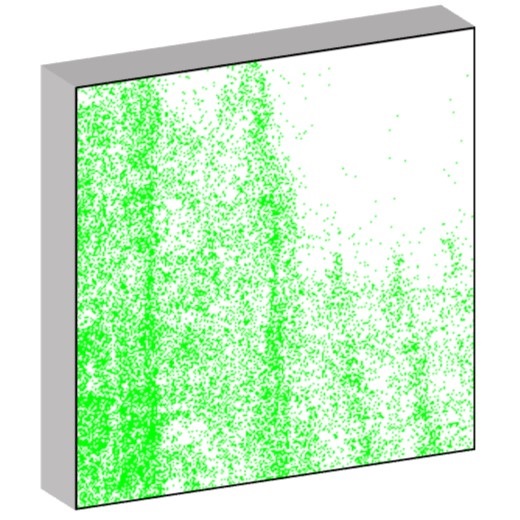}
			\caption{}
		\end{subfigure}
		\begin{subfigure}{.19\textwidth}
			\centering
			\includegraphics[width=\textwidth]{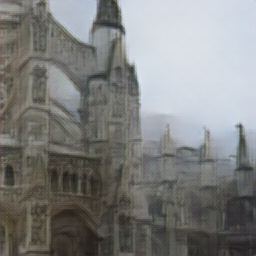}
			\caption{}
		\end{subfigure}
	\end{subfigure}
	\caption{Image reconstruction from solely SIFT coordinates. (a) Ground truth. (b)-(c) Inputs (binary map) and corresponding results. (d)-(e) Results of SLI with full SIFT features for comparison.}
	\label{fig:recon_location}
\end{figure}

\begin{table*}[t!]
	\caption{Quantitative comparison of the image reconstruction using solely SIFT descriptors or coordinates (binary map) over \texttt{CelebA-HQ}, \texttt{MD-NYU} and \texttt{ImageNet}. $^-$Lower is better. $^+$Higher is better.}
	\centering
	\label{tab:advanced}
	\begin{tabular}{ccccccccccccc}
		\toprule
		\multirow{2}{*}{Methods} & \multicolumn{4}{c}{\texttt{CelebA-HQ}} & \multicolumn{4}{c}{\texttt{MD-NYU}} & \multicolumn{4}{c}{\texttt{ImageNet}} \\
		
		\cmidrule(r){2-5} \cmidrule(r){6-9} \cmidrule(r){10-13}
		& FID$^-$ & SSIM$^+$ & PSNR$^+$ & PRM$^+$(\%)
		& FID$^-$ & SSIM$^+$ & PSNR$^+$ & PRM$^+$(\%)
		& FID$^-$ & SSIM$^+$ & PSNR$^+$ & PRM$^+$(\%)\\
		\midrule
		SLI-L
		& 181.0 & 0.372 & 13.70 & 13.52 & - & - & - & - & - & - & - & - \\
		SLI-R
		& 148.4 & 0.397 & 14.60 & 15.01 & 333.5 & 0.292 & 12.40 & 0.00 & 447.5 & 0.233 & 11.87 & 0.00 \\
		Coordinates
		& 122.4 & 0.449 & 13.88 & 14.43 & 243.8 & 0.238 & 12.33 & 1.19 & 440.3 & 0.234 & 11.80 & 0.00 \\
		SLI
		& \textbf{22.6} & \textbf{0.670} & \textbf{18.95} & \textbf{31.71} & \textbf{119.1} & \textbf{0.485} & \textbf{14.81} & \textbf{10.49} & \textbf{173.4} & \textbf{0.513} & \textbf{15.80} &
		\textbf{35.92}	\\
		\bottomrule
	\end{tabular}
\end{table*}

We then evaluate how much information can be reconstructed from \emph{solely} SIFT coordinates. Although SIFT coordinates are located in key regions of the image, they can only be represented as a binary map without specific image details. For the reconstruction result of using the binary map as the model input, a naive expectation is that the edge information can be well restored. Surprisingly, however, as shown in Fig. \ref{fig:recon_location}, the basic contours and contents of the objects in the image can be recovered, even though the lack of descriptors leads to blurred textures. The statistical results of the reconstructed images are also compiled into Table \ref{tab:advanced}. Compared with SLI-L and SLI-R, the reconstruction results from coordinates are slightly better. This validates the conclusion that the privacy leakage is not only through the descriptors, but also the coordinates.

The above results also imply that the privacy leakage problem is much less severe under \textbf{Scenario II} than the cases under \textbf{Scenario I} , especially when the adversary cannot access the coordinates.


\section{Conclusions}\label{sec:conclusion}

In this work, we have thoroughly investigated the privacy leakage problem of the widely-used SIFT features. We have first considered the \textbf{Scenario I}, where the adversary can fully access the SIFT features. We have proposed a deep generative model SLI for reconstructing the latent image from its SIFT features. The proposed model has been formed with two networks: a LBP reconstruction network, which aims to convert the SIFT into LBP features, and an image reconstruction network, which generates the reconstruction results by using the transformed LBP as a guidance. We then have considered the \textbf{Scenario II}, where the adversary can only access the partial SIFT features. We have designed landmark-based and reference-based methods for estimating SIFT coordinates from the descriptors. Experimental results have been provided to demonstrate the superiority of the proposed model SLI under these two scenarios. Our results also have suggested that the privacy leakage problem can be largely avoided under \textbf{Scenario II}, especially when the adversary cannot access the coordinates.

\ifCLASSOPTIONcaptionsoff
  \newpage
\fi



\bibliographystyle{IEEEtran}
\bibliography{ref}

%

%






\end{document}